\documentclass[journal]{IEEEtran}

\usepackage[T1]{fontenc}
\usepackage[utf8]{inputenc}  
\usepackage{lmodern}

\usepackage{amsmath,amssymb}
\usepackage{graphicx}
\usepackage{epstopdf}

\usepackage{booktabs}
\usepackage{siunitx}
\usepackage{multirow}
\usepackage{adjustbox}
\usepackage{threeparttable}
\usepackage{subcaption}

\usepackage{placeins}  
\usepackage{float}     

\usepackage[numbers,sort&compress]{natbib}

\usepackage{hyperref}
\hypersetup{
  colorlinks=true,
  linkcolor=blue,
  citecolor=blue,
  urlcolor=blue
}
\makeatletter

\renewcommand\section{\@startsection {section}{1}{\z@}%
  {-3.0ex plus -1ex minus -.2ex}%
  {1.8ex plus .2ex}%
  {\normalfont\Large\bfseries}}

\renewcommand\subsection{\@startsection{subsection}{2}{\z@}%
  {-2.0ex plus -.5ex minus -.2ex}%
  {1.0ex plus .2ex}%
  {\normalfont\large\bfseries}}
\makeatother

\makeatletter
\def\ps@IEEEtitlepagestyle{%
  \def\@oddfoot{\hfil\thepage\hfil}%
  \def\@evenfoot{\hfil\thepage\hfil}%
  \def\@oddhead{}%
  \def\@evenhead{}%
}
\def\ps@headings{%
  \def\@oddhead{}%
  \def\@evenhead{}%
  \def\@oddfoot{\hfil\thepage\hfil}%
  \def\@evenfoot{\hfil\thepage\hfil}%
}
\def\ps@plain{%
  \def\@oddhead{}%
  \def\@evenhead{}%
  \def\@oddfoot{\hfil\thepage\hfil}%
  \def\@evenfoot{\hfil\thepage\hfil}%
}
\makeatother

\usepackage{tikz}
\usepackage{xcolor}
\definecolor{neongreen}{RGB}{0,255,100}

\makeatletter
\AtBeginDocument{%
  \let\oldcitep\citep
  \renewcommand{\citep}[1]{%
    \oldcitep{#1}%
    \raisebox{-0.2ex}{\tikz{%
      \foreach \x in {1,...,\value{NAT@ctr}} {%
        \node[draw=neongreen, line width=2pt, fill=neongreen!15,
              inner sep=1pt, rectangle] at (\x*0.6,0) {\x};%
      }%
    }}%
  }%
}
\makeatother

\title{\bfseries Conditional Uncertainty-Aware Political Deepfake Detection with Stochastic Convolutional Neural Networks}
\author{\IEEEauthorblockN{\large  Rafael-Petruț Gardoș}
\\ \texttt{gardos.rafael@gmail.com}}

\begin{document}

\maketitle
\pagestyle{headings}

\def\baselinestretch{0.99}
\normalsize

\begin{abstract}
Recent advances in generative image models have enabled the creation of highly realistic
political deepfakes, posing serious risks to information integrity, public trust, and democratic
processes. While automated deepfake detectors are increasingly deployed in moderation and
investigative pipelines, most existing systems provide only point predictions and fail to indicate
when outputs are unreliable, being an operationally critical limitation in high-stakes political contexts.
This work investigated conditional, uncertainty-aware political deepfake detection using
stochastic convolutional neural networks within a strictly empirical, decision-oriented reliability
framework. Rather than framing uncertainty from a purely Bayesian or interpretive perspective,
uncertainty was evaluated through observable criteria, including calibration and its relationship
to prediction errors under global and confidence-conditioned evaluation regimes. A politically
focused binary image dataset was constructed via deterministic metadata-based filtering from a
large public real-synthetic corpus. Two pretrained CNN backbones, ResNet-18 and EfficientNet-
B4, were fully fine-tuned end-to-end for binary classification. Deterministic inference was
compared with stochastic procedures, including single-pass stochastic prediction, Monte Carlo
dropout with multiple forward passes, temperature scaling for calibration, and an ensemble-
based uncertainty surrogate as a non-Bayesian reference. Evaluation protocols were defined,
with the fake class treated as positive, ROC-AUC and thresholded confusion matrices reported,
and experiments conducted under controlled in-distribution settings with supplementary
generator-disjoint out-of-distribution analysis. The results showed that calibrated probabilistic
outputs and uncertainty estimates supported downstream decision-making by enabling risk-
aware moderation policies. A systematic confidence-band analysis further delineated when
uncertainty added operational value beyond predicted confidence, clarifying the practical scope
and limitations of uncertainty-aware deepfake detection in political contexts.
\end{abstract}

\section{Introduction}

Political deepfakes, synthetic images depicting public figures in fabricated or misleading contexts, represent a growing challenge for information ecosystems. Improvements in generative image models have enabled the creation of content that is increasingly difficult to distinguish from authentic imagery, even for expert human observers. In political contexts, such imagery can be used to fabricate events, misrepresent actions or statements, and erode trust in legitimate media sources.

Automated detection models are increasingly deployed to identify and triage potentially manipulated political imagery. Most detection systems are evaluated primarily using discriminative performance metrics such as accuracy or the area under the receiver operating characteristic curve (ROC-AUC) \cite{wang2024deepfake,kumar2025advances}. While these metrics quantify average separability between real and synthetic samples, they do not characterize the reliability of predicted probabilities. In high-stakes political settings, reliability is a critical operational property \cite{ gawlikowski2023survey,ovadia2019trust}. A detector that is overconfident on ambiguous or borderline inputs may be unsafe even if its average accuracy is high.

This motivates the integration of uncertainty-aware inference into political deepfake detection. However, uncertainty estimation is frequently discussed in abstract or interpretive terms. In this work, uncertainty is treated as an empirical signal whose value is determined by its observable behavior on held-out data. Specifically, uncertainty is evaluated according to two criteria: (i) probabilistic calibration, which measures the agreement between predicted confidence and empirical correctness, and (ii) uncertainty-error correlation, which measures whether uncertainty values are systematically higher for misclassified samples than for correctly classified ones.

A commonly used approach for uncertainty-aware inference in neural networks is Monte Carlo (MC) dropout, which activates dropout layers at test time and samples a predictive distribution via repeated stochastic forward passes \cite{gal2016dropout}. While MC dropout is often described as an approximate Bayesian method, its empirical behavior may reflect both posterior-like variability and noise-induced smoothing. Consequently, this work does not assume that MC dropout variance corresponds uniquely to epistemic uncertainty. Instead, uncertainty estimates are evaluated empirically and comparatively, alongside deterministic inference, single-pass stochastic inference, and post-hoc calibration baselines.

The central question addressed in this work is therefore not whether uncertainty estimates are theoretically Bayesian, but whether they are operationally useful. Formally, this work asks:

\begin{quote}
\textit{Under controlled in-distribution evaluation and under a limited generator-disjoint distribution shift, do uncertainty-aware inference procedures improve the reliability of political deepfake detector outputs, as measured by calibration and uncertainty-error correlation, relative to deterministic and post-hoc calibrated baselines?}
\end{quote}

This question is investigated using two convolutional neural network backbones that differ substantially in depth and representational capacity, evaluated under a unified experimental protocol with explicitly defined semantics and metrics.

\section{Related Work}

\subsection{Deepfake Detection in Political and Real-World Contexts}
Early deepfake detection methods relied on identifying low-level artifacts, physiological inconsistencies, or generator-specific cues embedded in synthetic imagery. While effective against early generative models, many such approaches degrade as synthesis quality improves and post-processing pipelines become more sophisticated. Contemporary detectors increasingly adopt deep convolutional architectures trained end-to-end \cite{yumlembam2025combined} to discriminate real from synthetic images based on learned representations rather than hand-crafted cues.

Despite strong performance under controlled benchmarks, deepfake detection in political contexts presents additional challenges \cite{wang2024deepfake,zhu2025antiforensic,lu2023video}. Political imagery frequently contains complex backgrounds, crowd scenes, non-frontal faces, varied lighting conditions, and substantial compression introduced by social media platforms. Moreover, political content is subject to rapid temporal drift as new events, figures, and visual styles emerge. These factors reduce margin separability and increase ambiguity even for authentic images, motivating evaluation beyond average accuracy. As a result, recent work has emphasized the need to assess detector behavior under uncertainty, rather than relying solely on discriminative metrics \cite{jin2025reliable,kose2025certainly}.

\subsection{Reliability and Calibration of Neural Network Predictions}
Probabilistic reliability has been extensively studied in the context of neural network classifiers. Calibration evaluates whether predicted probabilities align with empirical correctness frequencies \cite{guo2017calibration,niculescu2005predicting}, providing a complementary perspective to discrimination-focused metrics such as accuracy or ROC-AUC. Proper scoring rules, including negative log-likelihood and the Brier score, quantify probabilistic accuracy by penalizing overconfident incorrect predictions \cite{kuleshov2018accurate,he2023survey}, while summary metrics such as expected calibration error (ECE) aggregate discrepancies between confidence and accuracy across bins.

Importantly, calibration and discrimination capture distinct properties of a model. A classifier may achieve near-perfect ROC-AUC while remaining poorly calibrated \cite{thulasidasan2019mixup}, particularly when predictions are systematically overconfident. Conversely, post-hoc calibration methods can substantially alter confidence estimates without affecting ranking-based metrics. This distinction is especially relevant in high-stakes settings, where downstream decisions depend on the reliability of predicted probabilities rather than on ranking alone \cite{abdar2021uncertainty}. Consequently, recent work increasingly advocates reporting calibration-sensitive metrics alongside accuracy and AUC, particularly when model outputs are intended to support human-in-the-loop decision-making.

\subsection{Uncertainty Estimation via Stochastic Inference}
Stochastic inference techniques such as Monte Carlo dropout approximate predictive distributions by activating dropout layers at test time and sampling multiple stochastic forward passes \cite{gal2016dropout}. These methods are often motivated as approximations to Bayesian inference; however, their empirical behavior can also reflect noise-induced smoothing and regularization effects. As a result, predictive variance under dropout does not necessarily correspond to epistemic uncertainty in a strict sense \cite{kendall2017uncertainties,djupskas2025unreliable,wang2025aleatoric}.

Several studies have emphasized the importance of evaluating stochastic inference empirically rather than assuming a probabilistic interpretation \cite{huang2025conformal}. Comparisons against deterministic inference, single-pass stochastic inference, and non-stochastic calibration baselines are essential for disentangling the effects of injected noise from multi-sample aggregation \cite{buddenkotte2023ensembles}. From an operational perspective, the utility of uncertainty estimates lies not in their theoretical provenance but in their observable behavior: whether they improve calibration, reduce overconfidence, or correlate with prediction errors.

\subsection{Uncertainty, Selective Prediction, and Deepfake Detection}
Uncertainty-aware prediction has been explored in the broader context of selective classification, abstention mechanisms, and risk-coverage trade-offs, where models defer decisions on inputs deemed unreliable. In such settings, uncertainty is valuable insofar as it identifies cases likely to be incorrect \cite{hendrycks2016baseline,angelopoulos2021conformal}, enabling downstream policies such as escalation to human review or prioritization under limited resources.

In deepfake detection, relatively few works have evaluated uncertainty explicitly, and even fewer have examined uncertainty-error alignment in politically salient imagery. Existing approaches have considered ensembles or stochastic inference, but often report uncertainty qualitatively or assume a Bayesian interpretation without empirical validation. Systematic evaluation of whether uncertainty estimates meaningfully rank misclassified samples above correct ones remains limited, particularly under controlled political datasets.

The present work contributes to this line of research by evaluating uncertainty as an operational signal rather than an interpretive construct. By measuring calibration, proper scoring rules, and uncertainty-error correlation across multiple inference procedures and architectures, this study situates uncertainty-aware deepfake detection within a reliability-oriented evaluation framework aligned with real-world decision-making requirements.

\section{Methods}

\subsection{Dataset Construction and Filtering}

This study primarily evaluates political deepfake detection under controlled, in-distribution conditions. To this end, a politically focused binary image dataset is constructed from a large public real-synthetic image corpus, OpenFake, containing authentic photographs and AI-generated images produced by multiple generative models \cite{livernoche2025openfake}. The source corpus provides paired image data and structured metadata, including textual prompts, captions, generator identifiers, and ground-truth authenticity labels.

Formally, let
\begin{equation}
\mathcal{D} = \{(x_i, y_i)\}_{i=1}^{N}
\end{equation}
denote the resulting dataset, where $x_i$ is an image and $y_i \in \{0,1\}$ is its ground-truth label. The label convention is fixed throughout the manuscript such that $y_i = 1$ denotes a synthetic (fake) image and $y_i = 0$ denotes a real image.

\paragraph{Political filtering pipeline.}
Because the full source corpus spans diverse non-political content, a deterministic metadata-based filtering pipeline is applied to extract a subset relevant to political communication. The pipeline streams the source dataset iteratively and inspects available metadata fields, such as prompts, captions, textual descriptions, and categorical annotations, to identify references to political actors, institutions, events, or discourse. An image is retained if any metadata field contains at least one keyword from a predefined list:

\begin{verbatim}
POLITICAL_KEYWORDS = [
  "president", "prime minister",
  "election", "campaign", "senator",
  "congress", "parliament",
  "press conference", "speech", "rally",
  "biden", "trump", "harris",
  "trudeau", "sunak", "macron",
  "putin", "zelensky"
]
\end{verbatim}

This keyword list is fixed prior to dataset construction and applied uniformly to both real and synthetic samples. Importantly, the filtering step depends only on metadata and does not use image pixels or model predictions, preventing label leakage into subsequent learning or evaluation stages. All retained images are written to disk with structured filenames, and their associated metadata (label, keyword match, generator identifier, file path) is recorded in a CSV file to support reproducibility and downstream stratified analyses.

\paragraph{Dataset composition.}
After filtering, the final dataset contains $N = 4000$ images, balanced across classes with $2000$ real and $2000$ synthetic samples. 
The balanced design prevents trivial accuracy inflation due to class imbalance and ensures that calibration and uncertainty metrics are interpretable under symmetric class priors. The retained images span a range of political contexts, including campaign events, speeches, press conferences, and depictions of well-known public figures; however, coverage is determined by the distribution present in the source corpus and the keyword filter rather than by manual curation.

\paragraph{Train-validation-test partitioning.}
The dataset is partitioned at the image level into three disjoint subsets using a fixed random seed:
\begin{itemize}
  \item Training set: $2800$ images
  \item Validation set: $600$ images
  \item Test set: $600$ images
\end{itemize}

The dataset is partitioned using a random image-level split without stratification by political identity, event type, or generator. Generator labels may therefore appear in multiple splits; as a result, the in-distribution test evaluates matched-generator generalization rather than cross-generator generalization. Consequently, the same public figures, prompts, and generator families may appear in multiple splits, so the evaluation reflects \emph{in-distribution} performance under matched train-test conditions. In addition, a separate generator-disjoint OOD evaluation split is constructed by holding out specific generator families from training and reserving them exclusively for evaluation; political identities remain pooled.

\paragraph{Out-of-distribution (OOD) evaluation split.}
In addition to the in-distribution (ID) test split obtained via a random image-level partition, a separate out-of-distribution (OOD) evaluation split is constructed to assess \emph{generator-disjoint} generalization, following standard practice in distribution-shift evaluation for vision models \cite{hendrycks2019pretraining}. Concretely, the OOD split contains synthetic images produced by generator families not present in the training data (unseen generators at evaluation time), while political identities are not constrained to be disjoint and are therefore pooled across splits. The OOD split is used exclusively for evaluation (not for model selection), and all metrics follow the same evaluation semantics as ID (positive class: fake; ROC-AUC over $s(x)=p(y{=}1\mid x)$; accuracy and calibration at threshold $t{=}0.5$).

\paragraph{Scope and limitations.}
The dataset construction procedure is designed to support controlled evaluation of discriminative performance, calibration, and uncertainty-error relationships under matched train-test conditions. While the filtering pipeline yields a politically salient subset, it does not guarantee uniform coverage across all politicians, geopolitical regions, event types, imaging conditions, or generative methods. Accordingly, conclusions drawn from this dataset are restricted to the observed distribution and do not claim robustness to unseen identities, novel generators, or out-of-distribution political imagery.

\subsection{Model Architectures}

Two convolutional neural network backbones are evaluated: ResNet-18 and EfficientNet-B4 \cite{he2016resnet,tan2019efficientnet}. These architectures are selected to provide contrasting inductive biases, depth profiles, and parameterization regimes while remaining computationally feasible for repeated stochastic inference. ResNet-18 is a relatively shallow residual network characterized by uniform channel widths and additive skip connections, whereas EfficientNet-B4 employs compound scaling across depth, width, and input resolution, together with inverted residual blocks and squeeze-and-excitation mechanisms. Evaluating both models enables assessment of whether reliability and uncertainty behaviors are consistent across architectures with substantially different representational capacity and architectural design \cite{dosovitskiy2021vit}.

Each model defines a mapping
\begin{equation}
z(x) = g_{\phi}\big(h_{\theta}(x)\big),
\end{equation}
where $h_{\theta}(\cdot)$ denotes the convolutional feature extractor and $g_{\phi}(\cdot)$ denotes a learned classification head. The head consists of global average pooling followed by a linear projection to a single scalar logit. The predicted probability of the positive class is obtained via the logistic sigmoid
\begin{equation}
p(y{=}1 \mid x) = \sigma(z(x)) = \frac{1}{1 + \exp(-z(x))}.
\end{equation}
Throughout the manuscript, the positive class is fixed as $y=1$, corresponding to a synthetic (fake) image.

Both backbones are initialized with ImageNet-pretrained weights. Unless explicitly stated otherwise, all parameters $(\theta,\phi)$ are optimized jointly using end-to-end fine-tuning. This choice avoids assuming that head-only training suffices for comparability and reduces the risk of conflating uncertainty-related effects with representational underfitting caused by frozen feature extractors. Frozen-backbone and partial fine-tuning regimes are not explored in this study. 
All reported results correspond to full end-to-end fine-tuning, avoiding assumptions that head-only training suffices for comparability or uncertainty analysis.

\textbf{Dropout placement and stochasticity:} Dropout layers present in the original architectures are retained during fine-tuning. For uncertainty-aware inference, dropout is explicitly enabled at test time, inducing stochastic forward passes through both the backbone and the classification head. As a result, stochasticity affects intermediate feature representations as well as the final decision layer, ensuring that Monte Carlo samples reflect variability throughout the network rather than being confined to the classifier head alone.

\subsection{Preprocessing and Input Resolution}

All images are resized to a fixed spatial resolution prior to model input. Unless stated otherwise, a unified resolution of $380\times380$ pixels is used for both backbones. This choice deviates from canonical ResNet-18 preprocessing (typically $224\times224$), but is adopted to maintain a single preprocessing pipeline across architectures and to avoid introducing resolution-dependent confounds when comparing uncertainty behaviors. Input resolution is therefore treated as an experimental factor rather than a fixed assumption; its potential impact on discriminative performance and calibration is examined through targeted sensitivity analyses.

Pixel intensities are scaled to the interval $[0,1]$. Standard ImageNet mean-standard deviation normalization is not applied in the primary configuration. Omitting ImageNet normalization alters the input distribution relative to the pretrained feature extractors and may affect transfer performance or calibration. This deviation is treated explicitly as an experimental factor rather than an implicit assumption. 
To empirically justify the chosen preprocessing configuration, targeted ablations over input resolution and normalization strategy are conducted and reported in Section~\ref{sec:ablation_preproc}.

\subsection{Training Objective and Optimization}

Models are trained using the binary cross-entropy loss with logits. Given a batch of $n$ samples $\{(x_i,y_i)\}_{i=1}^{n}$, the loss is
\begin{equation}
\mathcal{L}_{\mathrm{BCE}} = -\frac{1}{n}\sum_{i=1}^{n}
\left[
y_i \log \sigma(z_i) + (1-y_i)\log\big(1-\sigma(z_i)\big)
\right],
\end{equation}
where $z_i = z(x_i)$ and $y_i \in \{0,1\}$. Optimization is performed using a fixed optimizer configuration across all models and inference variants to prevent procedural differences from confounding uncertainty comparisons. All random seeds are fixed per experiment, and training hyperparameters (learning rate, batch size, number of epochs, and weight decay) are held constant across backbones unless explicitly stated.

\subsection{Reproducibility and Experimental Details}

All experiments were conducted using a fixed and fully specified training and
evaluation protocol to ensure reproducibility. Models were implemented in
PyTorch~(v2.0.1) using torchvision~(v0.15.2). Training and inference were executed
on a single NVIDIA RTX~3090 GPU (24~GB VRAM) with CUDA~11.8; CPU preprocessing used
an Intel Xeon Silver 4210 processor. The operating system was Ubuntu~22.04~LTS.

\paragraph{Optimization.}
All models were optimized using the Adam optimizer with default momentum
parameters $(\beta_1{=}0.9,\ \beta_2{=}0.999)$. Unless otherwise stated, the
learning rate was set to $1\times10^{-4}$ for the classification head and
$1\times10^{-5}$ for backbone parameters during full end-to-end fine-tuning.
Weight decay was fixed at $1\times10^{-4}$. No gradient clipping was applied.

\paragraph{Training schedule.}
Models were trained for a maximum of 30 epochs with early stopping based on
validation negative log-likelihood, using a patience of 5 epochs. Early stopping
was used solely for checkpoint selection; the test set was not accessed during
training, calibration, or model selection. The model checkpoint with the best
validation NLL was selected for final evaluation on the test set. Learning rates
were held constant throughout training; no cosine decay or warm-up schedules
were used in the primary experiments.

\paragraph{Batching and data loading.}
All experiments used a batch size of 32 images. Data loading employed four worker
processes with pinned memory enabled. No class reweighting, oversampling, or
cost-sensitive training strategies were applied, as all dataset splits were
class-balanced by construction.

\paragraph{Dropout and stochastic inference.}
Dropout layers present in the pretrained architectures were retained during
fine-tuning. For Monte Carlo dropout inference, dropout was explicitly enabled
at test time. Unless otherwise stated, MC dropout results correspond to
$T{=}20$ stochastic forward passes. Single-pass stochastic inference ($T{=}1$)
uses the same dropout configuration but omits Monte Carlo averaging, isolating
the effect of test-time stochasticity from multi-sample aggregation.

\paragraph{Calibration and ensembles.}
Temperature scaling parameters were fitted exclusively on the validation set by
minimizing negative log-likelihood and then applied unchanged to the test set.
Ensemble results correspond to $K{=}5$ independently trained models with
different random initializations and identical architectures, preprocessing,
and optimization hyperparameters.

\paragraph{Random seeds.}
To control stochasticity, all experiments were conducted with fixed random seeds
for Python, NumPy, and PyTorch. The primary results reported in this paper use
seed~42. To assess robustness, key metrics (accuracy, ROC-AUC, ECE, and Brier
score) were additionally verified across three seeds $\{21, 42, 84\}$, yielding
qualitatively consistent trends. All tables and figures report results from the
primary seed, with metric uncertainty quantified via bootstrap confidence
intervals.

\paragraph{Evaluation protocol.}
All metrics are computed on held-out test sets that are not used during training,
calibration, or model selection. ROC-AUC is computed using probabilistic scores
$s(x)=p(y{=}1\mid x)$ without thresholding, while accuracy and confusion matrices
use a fixed decision threshold of $0.5$. Bootstrap confidence intervals are
computed using 1{,}000 resamples unless otherwise stated.

\paragraph{Code and data availability.}
All scripts required to reproduce preprocessing, training, inference, and
evaluation, including random seed configuration and figure generation, will be
released publicly upon acceptance. Dataset construction relies exclusively on
publicly available sources and deterministic metadata-based filtering procedures
defined in the Methods section.

\subsection{Inference Procedures and Experimental Controls}

To disentangle the effects of stochastic regularization from multi-sample Bayesian-style averaging, multiple inference procedures are evaluated under otherwise identical conditions.

\subsubsection{Deterministic Inference}

In deterministic inference, dropout layers are disabled at test time and a single forward pass is performed:
\begin{equation}
\hat{p}_{\mathrm{det}}(x) = \sigma(z(x)).
\end{equation}

\subsubsection{Single-Pass Stochastic Inference ($T=1$)}

To isolate the effect of test-time stochasticity without Monte Carlo averaging, a single stochastic forward pass is performed with dropout enabled:
\begin{equation}
\hat{p}_{T=1}(x) = \sigma(z_{1}(x)),
\end{equation}
where $z_{1}(x)$ corresponds to one dropout-sampled subnetwork. This condition controls for noise-induced smoothing independently of multi-sample aggregation.

\subsubsection{Monte Carlo Dropout Inference ($T>1$)}

With dropout enabled at test time, $T$ stochastic forward passes are drawn:
\begin{equation}
\{\hat{p}_{t}(x)\}_{t=1}^{T}, \quad \hat{p}_{t}(x)=\sigma(z_{t}(x)).
\end{equation}
The predictive mean is
\begin{equation}
\hat{\mu}(x) = \frac{1}{T}\sum_{t=1}^{T}\hat{p}_{t}(x),
\end{equation}
and the predictive variance is
\begin{equation}
\hat{\sigma}^2(x)=\frac{1}{T}\sum_{t=1}^{T}\hat{p}_{t}(x)^2-\hat{\mu}(x)^2.
\end{equation}
Predictive entropy is computed as
\begin{equation}
H(x)=-\hat{\mu}(x)\log \hat{\mu}(x)-(1-\hat{\mu}(x))\log(1-\hat{\mu}(x)).
\end{equation}

The $T=1$ and $T>1$ settings are evaluated side by side to determine whether observed calibration changes arise from stochastic regularization alone or from multi-sample predictive aggregation.

\subsubsection{Post-hoc Calibration and Ensemble Baselines}

Temperature scaling is included as a post-hoc calibration baseline. A scalar temperature parameter $\tau>0$ is fitted on the validation set by minimizing negative log-likelihood, and calibrated probabilities are computed as
\begin{equation}
\hat{p}_{\mathrm{temp}}(x)=\sigma\!\left(\frac{z(x)}{\tau}\right).
\end{equation}
Because temperature scaling preserves score ordering, it does not affect ranking-based metrics such as ROC-AUC, providing a control condition that modifies calibration without introducing stochasticity.

In practice, $\tau$ is estimated by minimizing validation-set NLL with respect to the scalar parameter $\tau$ using the deterministic logits $z(x)$. To enforce $\tau>0$, optimization is performed over $\alpha \in \mathbb{R}$ with $\tau=\exp(\alpha)$. Calibration is applied to logits rather than probabilities, i.e., $\hat{p}_{\mathrm{temp}}(x)=\sigma(z(x)/\tau)$.

An ensemble-based reference baseline is included by training $K$ independent models with different random seeds and shuffling orders under identical hyperparameters, and averaging their predictive probabilities at test time \cite{lakshminarayanan2017deepensembles}. Because this baseline does not rely on test-time dropout, it is treated as a non-dropout uncertainty mechanism.

\subsection{Evaluation Semantics and Thresholding}

All evaluation conventions are defined explicitly to prevent ambiguity. The positive class is fixed as $y=1$ (fake). Ranking metrics such as ROC-AUC are computed using scores
\begin{equation}
s(x)=p(y{=}1\mid x),
\end{equation}
while fixed-threshold metrics (accuracy and confusion matrices) use the decision rule
\begin{equation}
\hat{y}=\mathbb{I}[s(x)\ge 0.5].
\end{equation}
Unless otherwise stated, ROC curves and confusion matrices are computed on the same held-out test split.

\subsection{Performance, Calibration, and Uncertainty Metrics}

Accuracy is defined as
\begin{equation}
\mathrm{Acc}=\frac{1}{n}\sum_{i=1}^{n}\mathbb{I}[\hat{y}_i=y_i].
\end{equation}
Negative log-likelihood and the Brier score are computed as
\begin{equation}
\mathrm{NLL}=-\frac{1}{n}\sum_{i=1}^{n}\left[y_i\log\hat{p}_i+(1-y_i)\log(1-\hat{p}_i)\right],
\end{equation}
\begin{equation}
\mathrm{Brier}=\frac{1}{n}\sum_{i=1}^{n}(\hat{p}_i-y_i)^2.
\end{equation}

Expected calibration error (ECE) is computed by binning predictions by confidence and measuring the weighted discrepancy between empirical accuracy and mean confidence across bins \cite{guo2017calibration}. Calibration curves are reported descriptively; quantitative comparisons rely on ECE and proper scoring rules.

\subsection{Uncertainty-Error Correlation Metrics}

To assess whether uncertainty estimates are operationally informative, uncertainty-error alignment is quantified. For inference method $m$, the error indicator is
\begin{equation}
e_i^{(m)}=\mathbb{I}[\hat{y}_i^{(m)}\neq y_i],
\end{equation}
and an uncertainty score $u_i^{(m)}$ (e.g., predictive variance or entropy) is used as a ranking function. The area under the ROC curve for error detection is computed by treating $e_i^{(m)}$ as the positive label. High AUROC indicates that uncertainty systematically ranks misclassified samples above correctly classified ones.

\textbf{Interpretability scope:}
Throughout this work, uncertainty is interpreted narrowly as a decision-level signal. No claims are made regarding feature-level, causal, or mechanistic interpretability. All conclusions concerning interpretability refer exclusively to the operational utility of calibrated probabilities and uncertainty scores for downstream decision-making.

\section{Results}

\subsection{Confusion matrices at a fixed decision threshold ($t=0.5$)}
\begin{figure*}[t]
\centering
\begin{subfigure}{0.48\linewidth}
  \includegraphics[width=\linewidth]{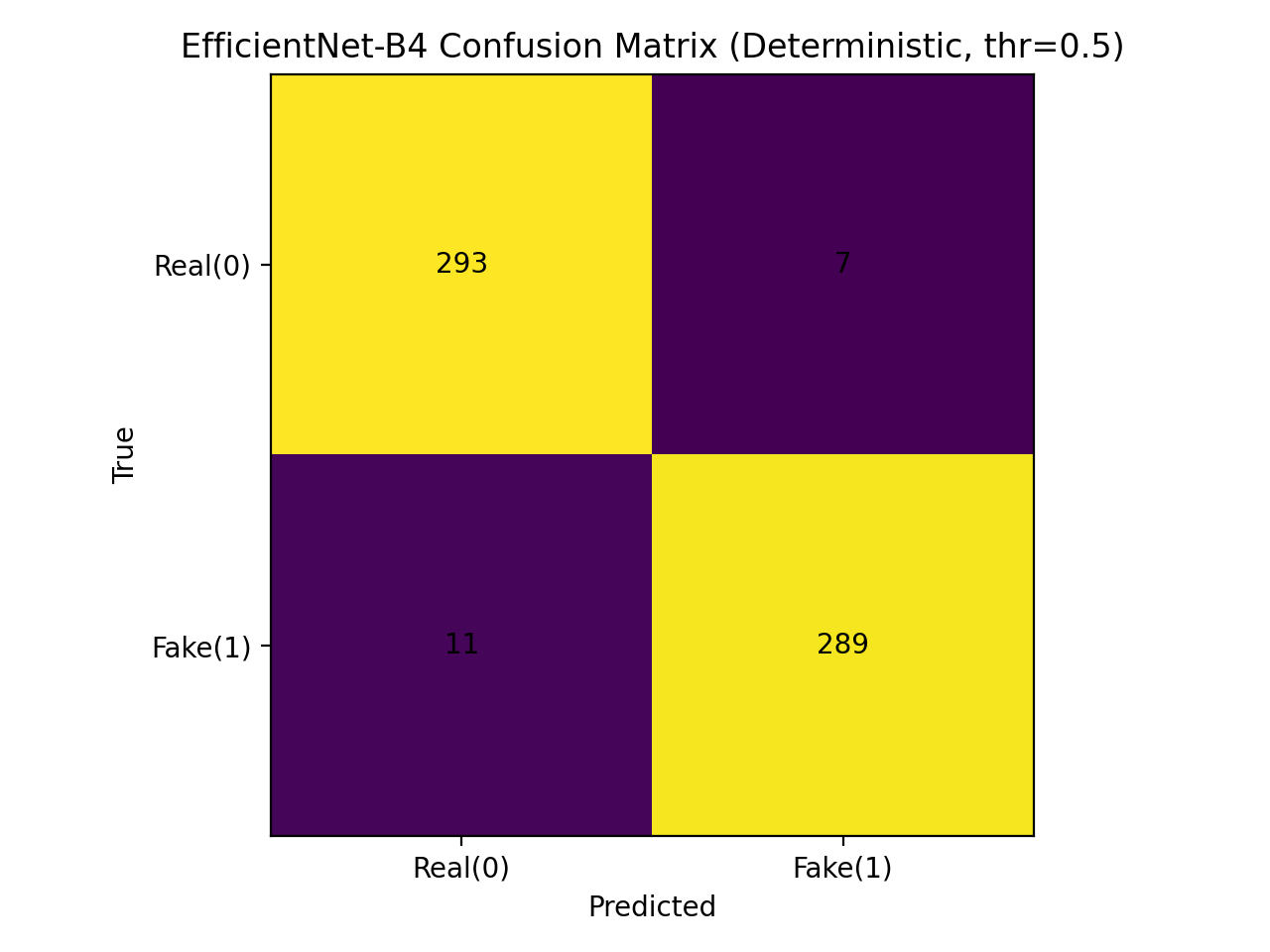}
  \caption{EfficientNet-B4}
\end{subfigure}\hfill
\begin{subfigure}{0.48\linewidth}
  \includegraphics[width=\linewidth]{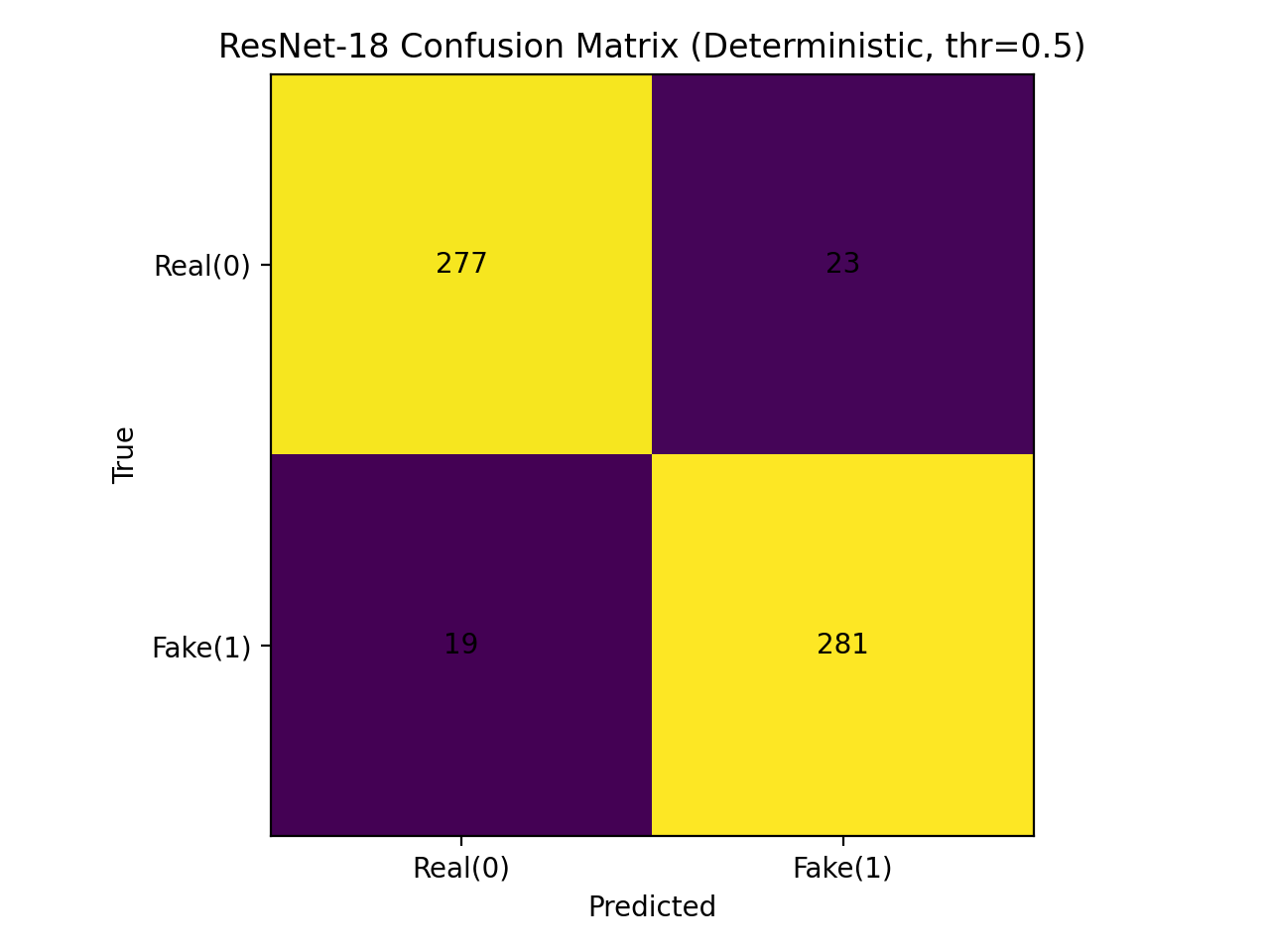}
  \caption{ResNet-18}
\end{subfigure}
\caption{Deterministic confusion matrices at $t=0.5$ (positive class: fake).}
\label{fig:cm_pair}
\vspace{0.6\baselineskip}
\end{figure*}

This subsection reports deterministic confusion matrices , reported in Figure ~\ref{fig:cm_pair}, computed on the held-out test split ($n=600$) under a fixed operating point $t=0.5$, with fake as the positive class ($y=1$). EfficientNet-B4 yields $289$ true positives, $293$ true negatives, $7$ false positives, and $11$ false negatives. ResNet-18 yields $281$ true positives, $277$ true negatives, $23$ false positives, and $19$ false negatives. In both cases, the counts sum to $600$, ensuring consistency between the reported confusion matrices and the stated test-set size.

\subsection{Discrimination performance: ROC curves and ROC-AUC}

\begin{figure*}[t]
\centering
\begin{subfigure}{0.48\linewidth}
  \includegraphics[width=\linewidth]{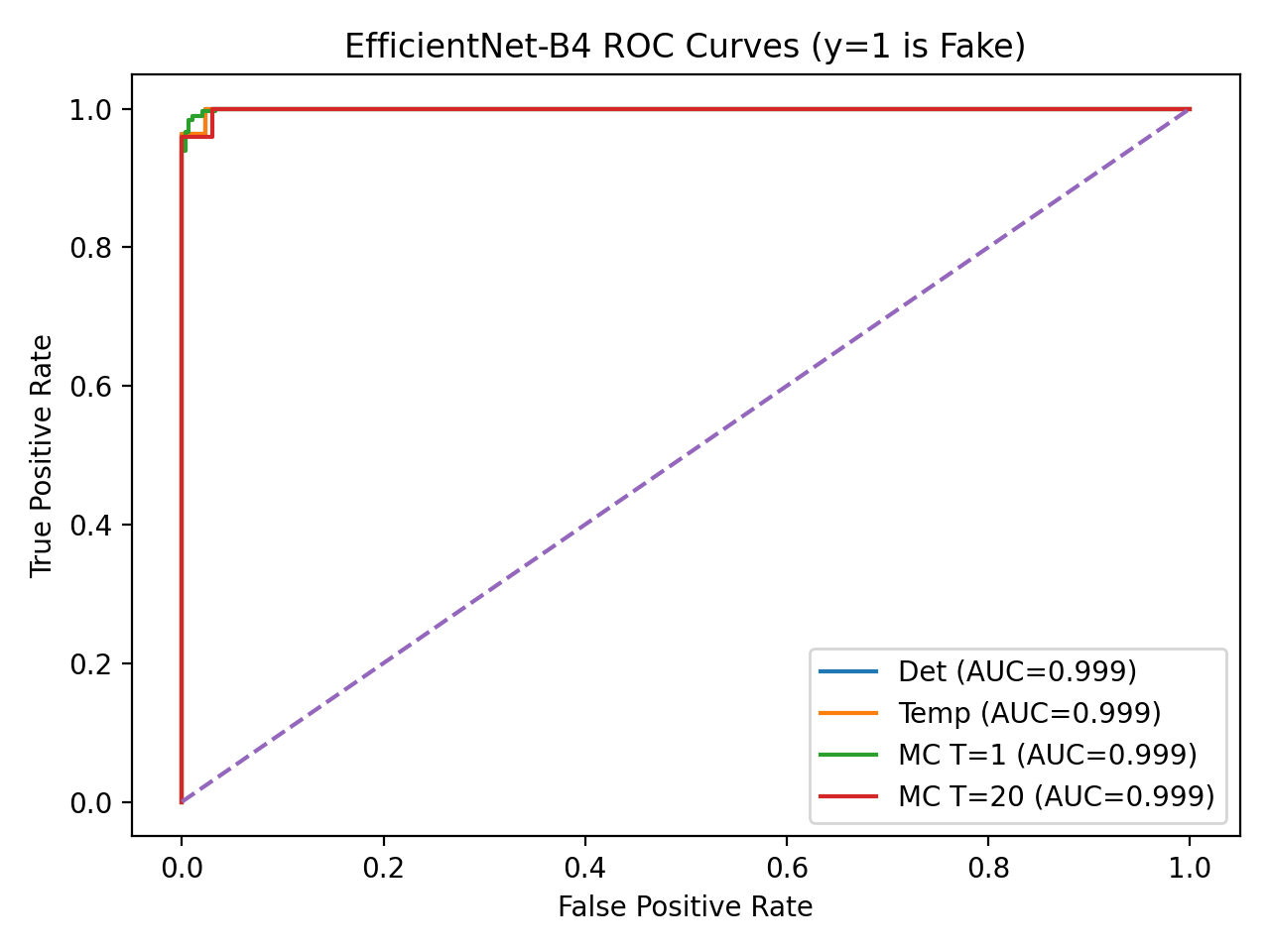}
  \caption{EfficientNet-B4}
\end{subfigure}\hfill
\begin{subfigure}{0.48\linewidth}
  \includegraphics[width=\linewidth]{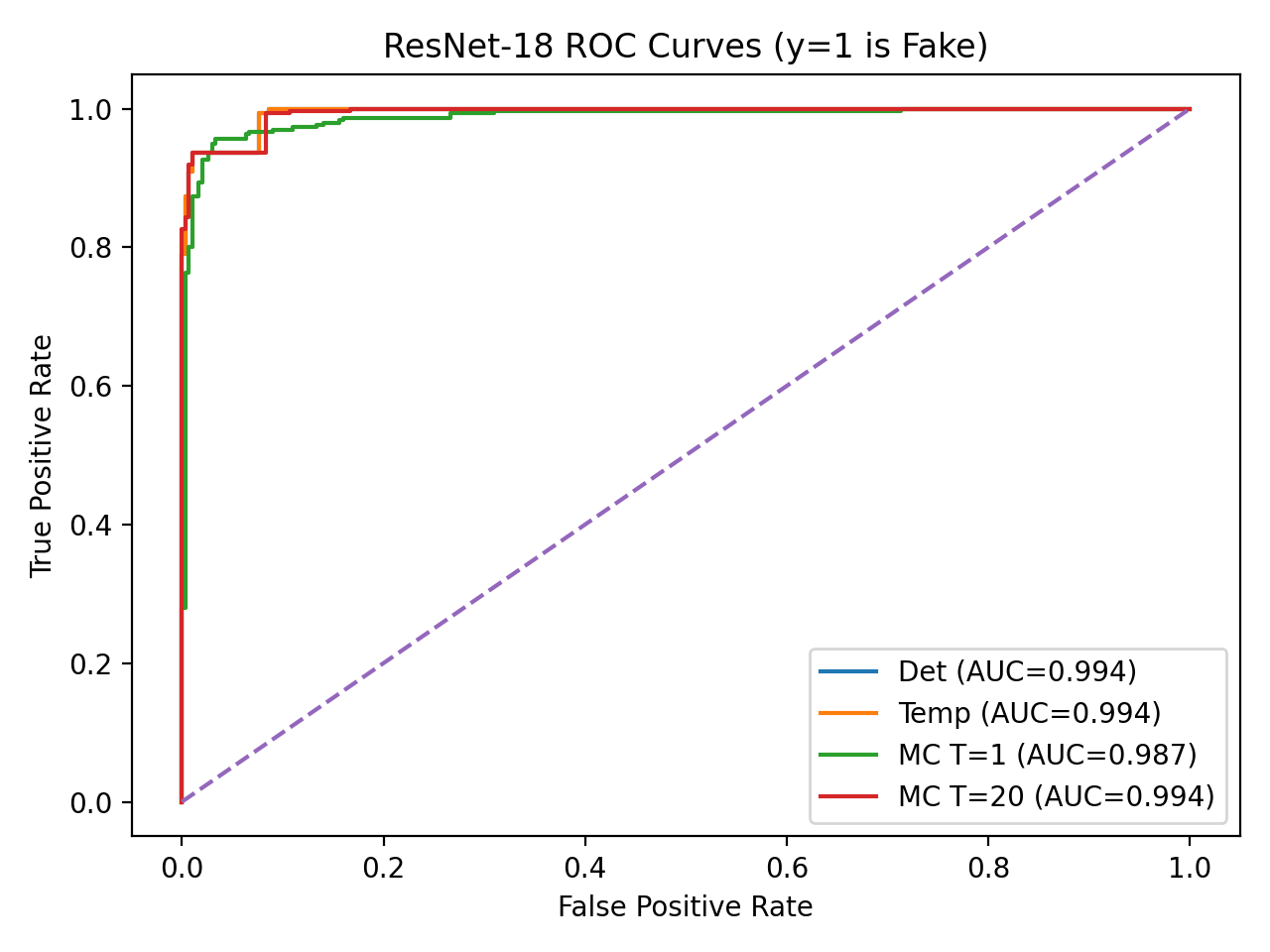}
  \caption{ResNet-18}
\end{subfigure}
\caption{ROC curves for $s(x)=p(y{=}1\mid x)$ (fake is positive).}
\label{fig:roc_pair}
\vspace{0.6\baselineskip}
\end{figure*}

Figure~\ref{fig:roc_pair} reports ROC curves obtained by varying the decision threshold over the score $s(x)=p(y{=}1\mid x)$. Under deterministic inference, the ROC curves lie close to the upper-left region; the corresponding ROC-AUC values are reported in Subsection~\ref{subsec:metrics_summary}. As a diagnostic principle, ROC-AUC below $0.5$ would indicate systematic misranking (e.g., score direction or label convention mismatch) rather than merely weak separability; the observed ROC curves and associated AUC values do not exhibit that behavior.

\subsection{Performance and reliability summary across inference procedures}
\label{subsec:metrics_summary}

This subsection consolidates accuracy, discrimination, and probabilistic reliability for the full set of inference procedures: deterministic inference, temperature scaling, single stochastic forward pass ($T{=}1$), MC dropout mean prediction ($T{=}20$), and an ensemble surrogate ($K{=}5$). Results are reported separately for each backbone to avoid conflating architectural effects. I report three descriptive observations.

Across both backbones, ROC-AUC values vary only slightly across procedures (Tables~\ref{tab:eff_metrics} and \ref{tab:res_metrics}). Second, calibration-sensitive metrics (ECE, Brier, NLL) differ across procedures, even when ROC-AUC is similar. Third, calibration-sensitive metrics differ across backbones and inference procedures (Tables~\ref{tab:eff_metrics} and \ref{tab:res_metrics}).

\begin{table*}[t]
\centering
\caption{EfficientNet-B4: performance and reliability across inference procedures (test set).}
\label{tab:eff_metrics}
\sisetup{round-mode=places,round-precision=4}
\begin{adjustbox}{max width=\linewidth}
\begin{tabular}{l S[table-format=1.4] S[table-format=1.4] S[table-format=1.4] S[table-format=1.4] S[table-format=1.4] S[table-format=3.3] S[table-format=3.3] S[table-format=3.3] S[table-format=3.3]}
\toprule
{Method} & {Acc} & {AUC} & {ECE} & {Brier} & {NLL} & {TP} & {FP} & {TN} & {FN}\\
\midrule
Deterministic & 0.970000 & 0.999144 & 0.067400 & 0.011426 & 0.075421 & 289 & 7 & 293 & 11\\
TempScaling & 0.970000 & 0.999144 & 0.148287 & 0.027746 & 0.165557 & 289 & 7 & 293 & 11\\
MC Dropout T=1 & 0.986667 & 0.999489 & 0.054927 & 0.012480 & 0.077868 & 295 & 3 & 297 & 5\\
MC Dropout T=20 mean & 0.965000 & 0.998800 & 0.071846 & 0.013548 & 0.081564 & 288 & 9 & 291 & 12\\
Ensemble surrogate (K=5) & 0.985000 & 0.999544 & 0.064007 & 0.012121 & 0.080416 & 294 & 3 & 297 & 6\\
\bottomrule
\end{tabular}
\end{adjustbox}
\end{table*}

\begin{table*}[t]
\centering
\caption{ResNet-18: performance and reliability across inference procedures (test set).}
\label{tab:res_metrics}
\sisetup{round-mode=places,round-precision=4}
\begin{tabular}{l S[table-format=1.4] S[table-format=1.4] S[table-format=1.4] S[table-format=1.4] S[table-format=1.4] S[table-format=3.3] S[table-format=3.3] S[table-format=3.3] S[table-format=3.3]}
\toprule
{Method} & {Acc} & {AUC} & {ECE} & {Brier} & {NLL} & {TP} & {FP} & {TN} & {FN}\\
\midrule
Deterministic & 0.930000 & 0.994289 & 0.137991 & 0.041077 & 0.180961 & 281 & 23 & 277 & 19\\
TempScaling & 0.930000 & 0.994289 & 0.254465 & 0.081004 & 0.319482 & 281 & 23 & 277 & 19\\
MC Dropout T=1 & 0.953333 & 0.987056 & 0.090221 & 0.045438 & 0.191890 & 287 & 15 & 285 & 13\\
MC Dropout T=20 mean & 0.926667 & 0.993633 & 0.139221 & 0.042714 & 0.185327 & 281 & 25 & 275 & 19\\
Ensemble surrogate (K=5) & 0.946667 & 0.994200 & 0.114771 & 0.045079 & 0.193885 & 287 & 19 & 281 & 13\\
\bottomrule
\end{tabular}
\end{table*}

\subsection{Score distribution diagnostics (deterministic)}

\begin{figure*}[t]
\centering
\begin{subfigure}{0.48\linewidth}
  \includegraphics[width=\linewidth]{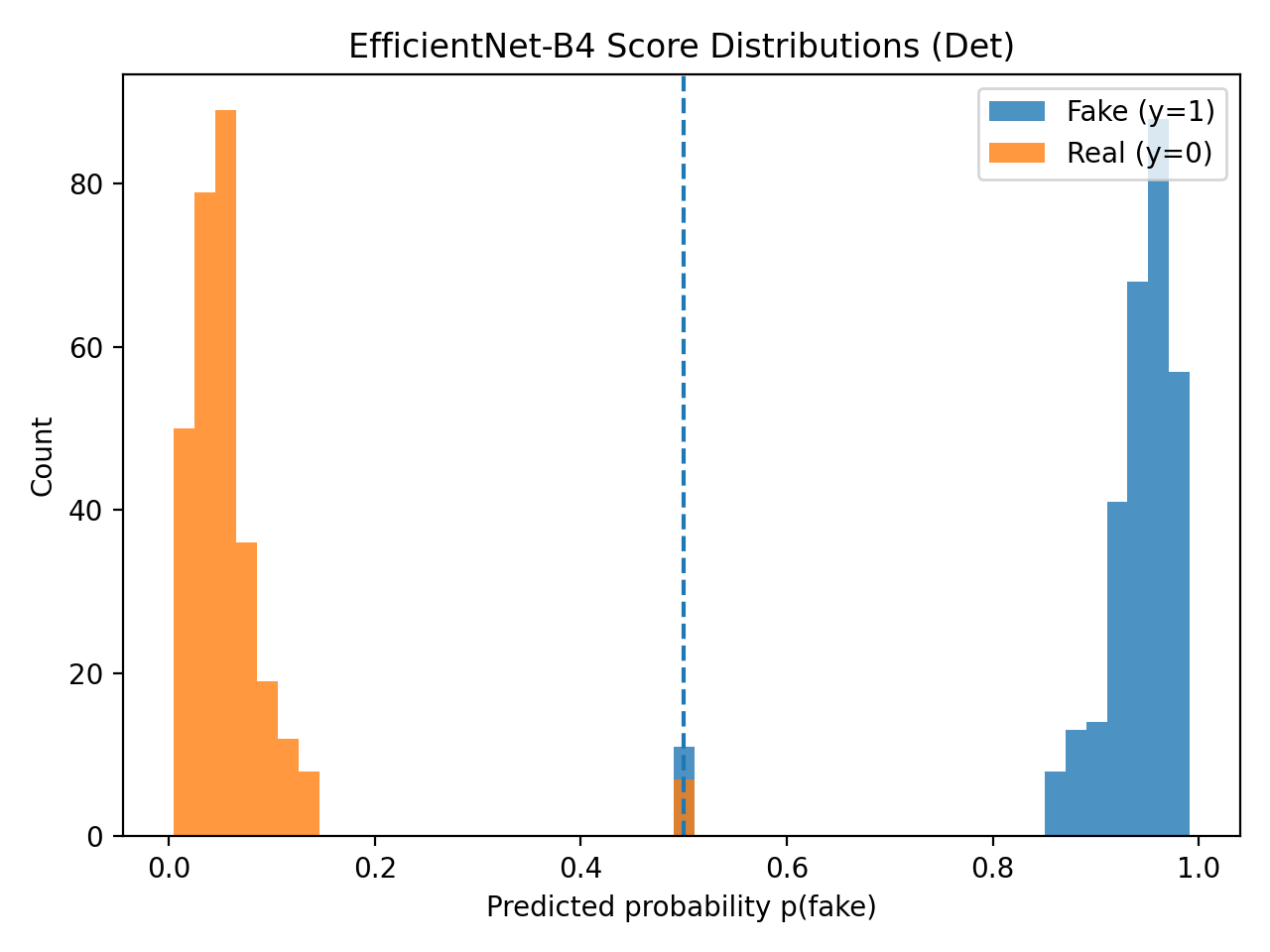}
  \caption{EfficientNet-B4}
\end{subfigure}\hfill
\begin{subfigure}{0.48\linewidth}
  \includegraphics[width=\linewidth]{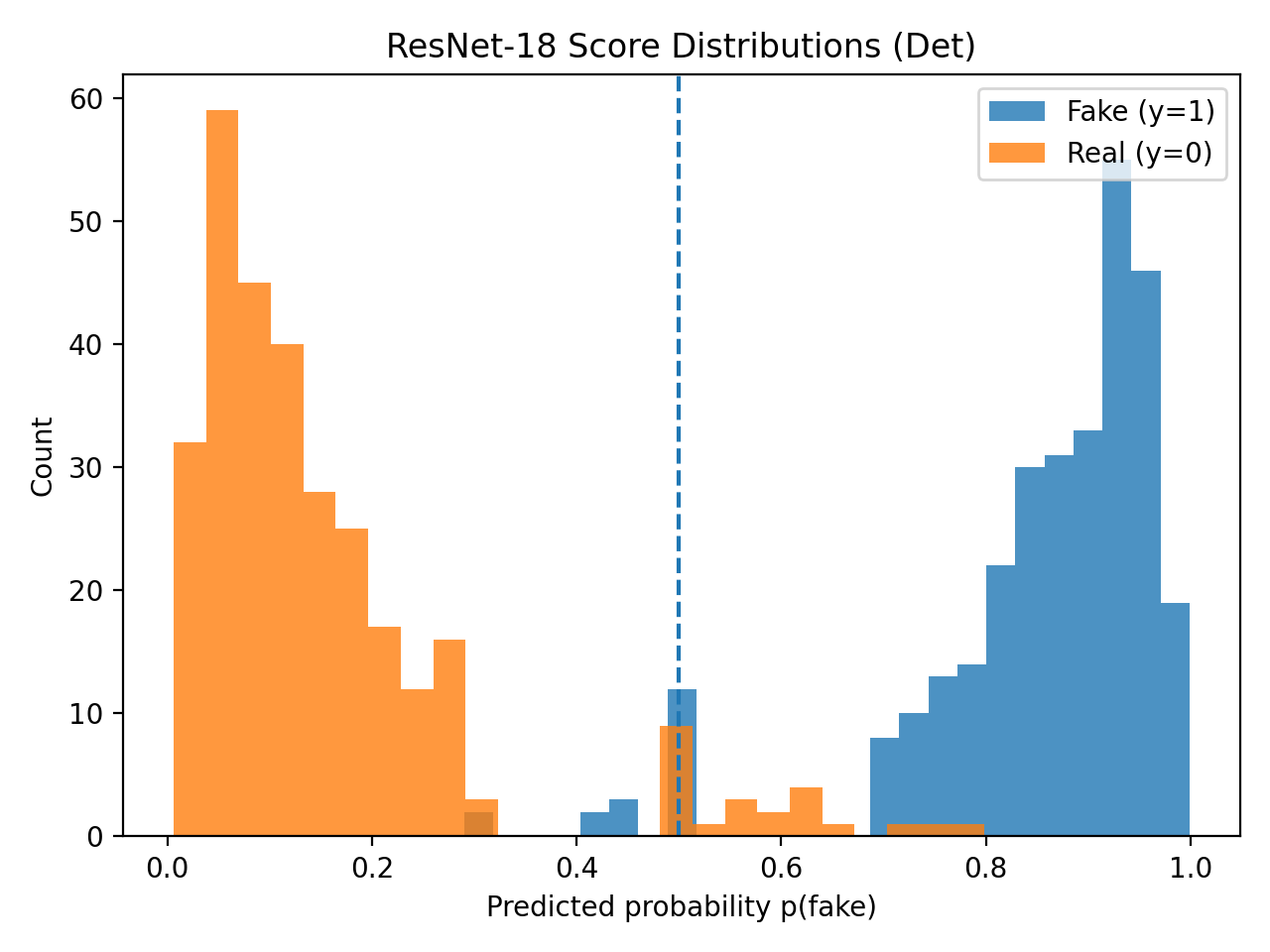}
  \caption{ResNet-18}
\end{subfigure}
\caption{Deterministic score distributions $s(x)=p(y{=}1\mid x)$ by class.}
\label{fig:score_hist_pair}
\end{figure*}

Figure~\ref{fig:score_hist_pair} visualizes the distribution of deterministic scores $s(x)=p(y{=}1\mid x)$ by class. For both backbones, real samples concentrate at low predicted fake probability while synthetic samples concentrate at high predicted fake probability, which align with the high ROC-AUC values reported earlier. The histograms provide a complementary view of score separation around the fixed threshold ($t=0.5$).

\subsection{Calibration baseline: deterministic vs.\ temperature scaling}

\begin{figure*}[t]
\centering
\begin{subfigure}{0.48\linewidth}
  \includegraphics[width=\linewidth]{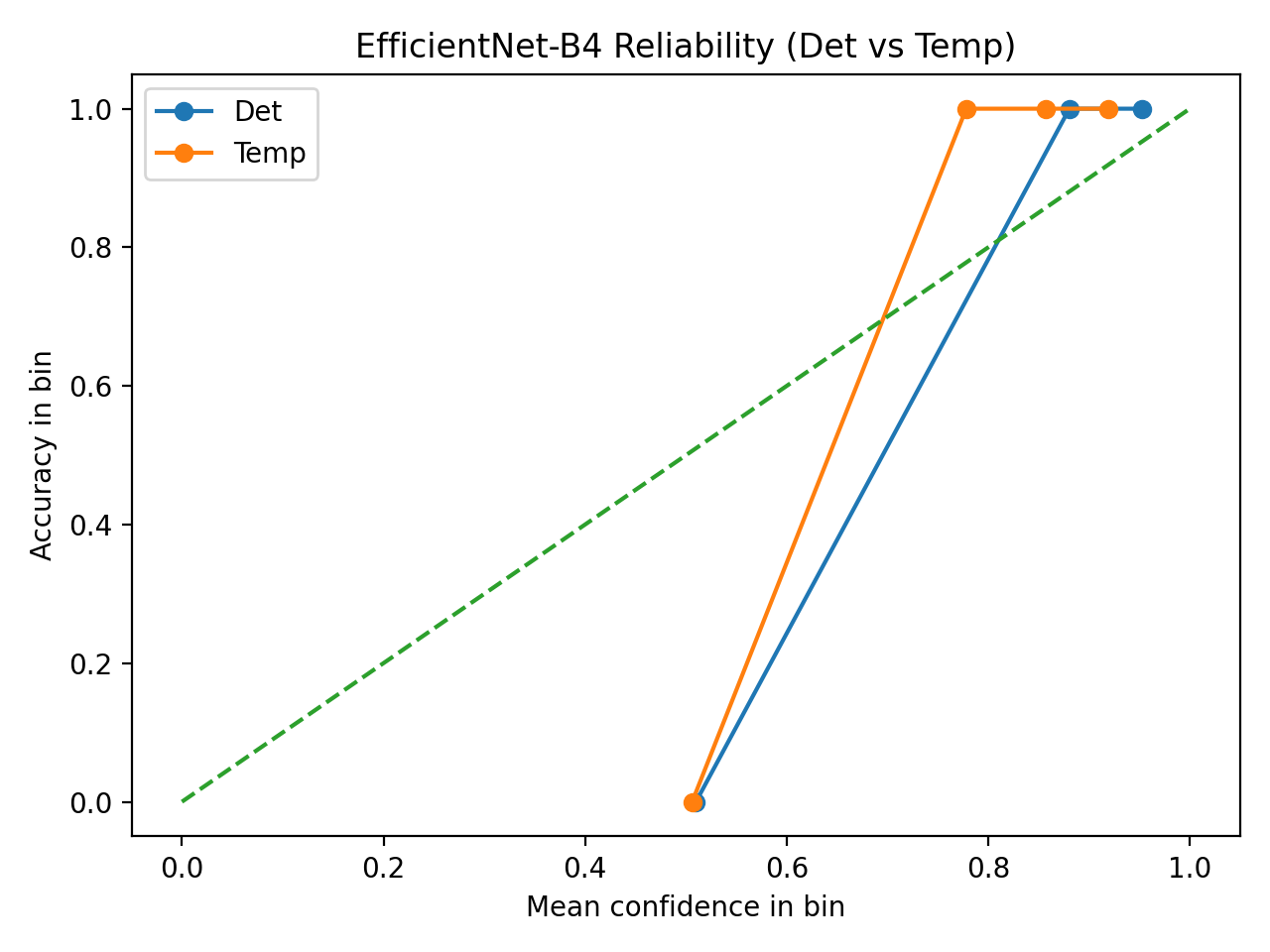}
  \caption{EfficientNet-B4}
\end{subfigure}\hfill
\begin{subfigure}{0.48\linewidth}
  \includegraphics[width=\linewidth]{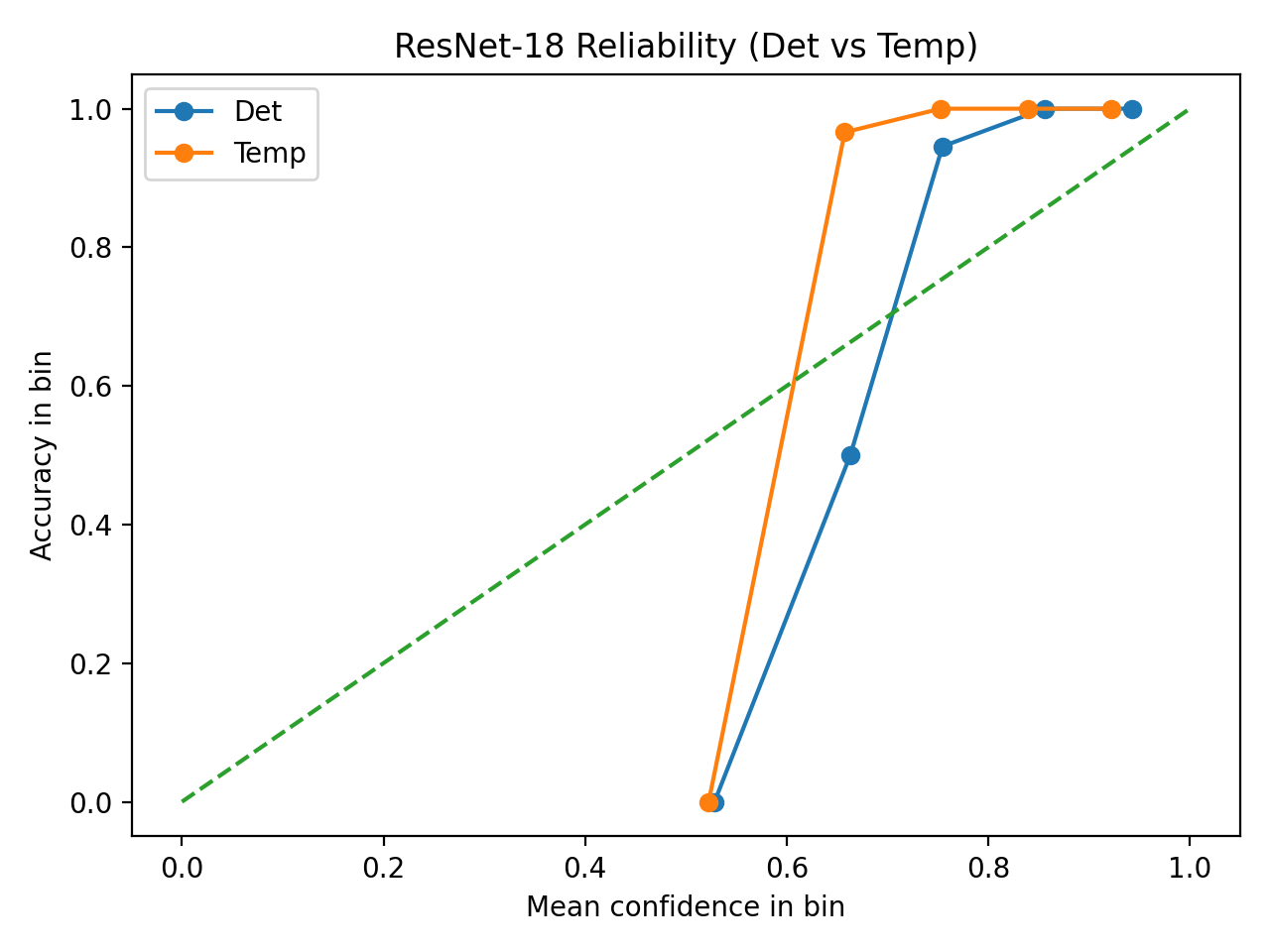}
  \caption{ResNet-18}
\end{subfigure}
\caption{Reliability diagrams: deterministic vs.\ temperature scaling.}
\label{fig:rel_temp_pair}
\end{figure*}

Figure~\ref{fig:rel_temp_pair} compares reliability diagrams for deterministic inference and temperature scaling. Because temperature scaling applies a monotone transformation of logits, it does not alter ranking-based metrics such as ROC-AUC, but it can substantially change calibration-sensitive metrics (ECE, NLL, Brier), as quantified in Tables~\ref{tab:eff_metrics} and \ref{tab:res_metrics}. Empirically, the calibration response to temperature scaling differs across backbones and depends on the fitted temperature.

\subsection{MC dropout mean calibration: deterministic vs.\ MC ($T=20$)}

\begin{figure*}[t]
\centering
\begin{subfigure}{0.48\linewidth}
  \includegraphics[width=\linewidth]{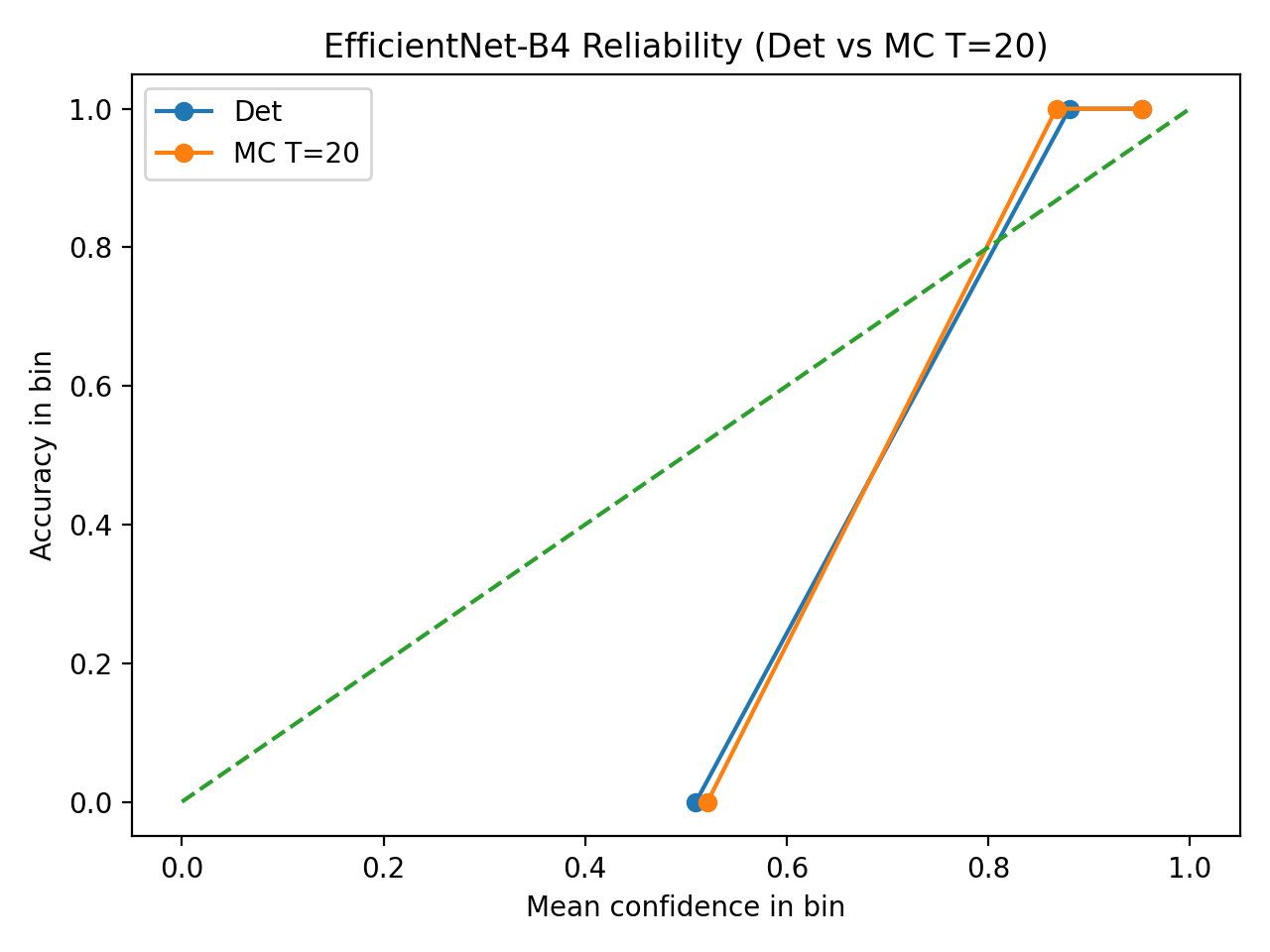}
  \caption{EfficientNet-B4}
\end{subfigure}\hfill
\begin{subfigure}{0.48\linewidth}
  \includegraphics[width=\linewidth]{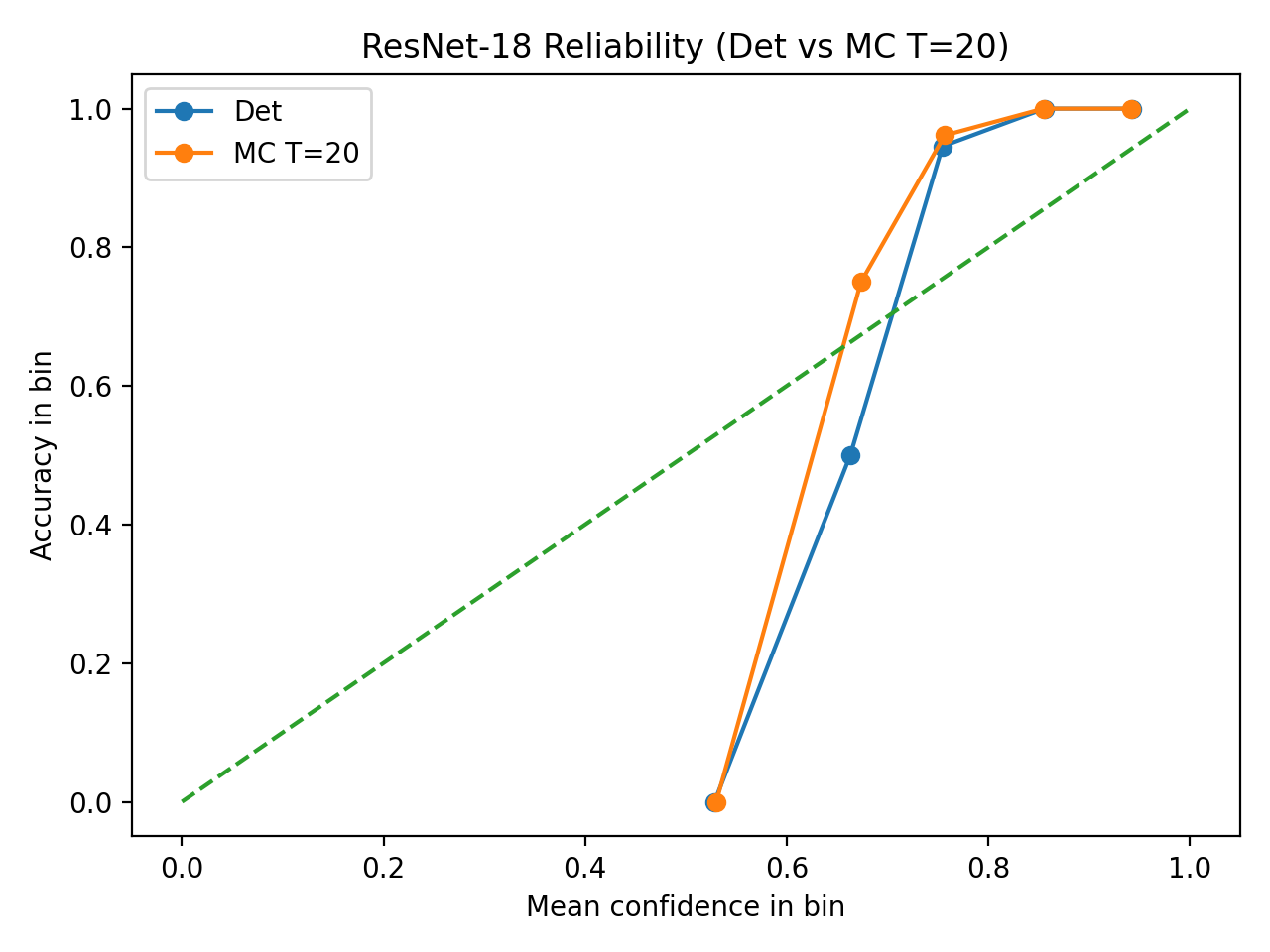}
  \caption{ResNet-18}
\end{subfigure}
\caption{Reliability diagrams: deterministic vs.\ MC dropout mean ($T=20$).}
\label{fig:rel_mc20_pair}
\end{figure*}

Figure~\ref{fig:rel_mc20_pair} compares deterministic reliability against MC dropout mean predictions with $T=20$ stochastic forward passes. Averaging stochastic predictions can change both the sharpness of predicted probabilities and the distribution of confidence values, which is associated with changes in calibration measures. The resulting ECE and proper scoring rules are reported in Tables~\ref{tab:eff_metrics} and \ref{tab:res_metrics}. Calibration metrics are therefore additionally reported as a function of $T$ in the $T$-sensitivity analysis below.

\subsection{Single-pass stochastic inference as a control condition ($T=1$)}

Single-pass stochastic inference ($T=1$) activates dropout at test time but does not perform multi-sample averaging. Consequently, it isolates the effect of injecting stochasticity at inference from the effect of Monte Carlo averaging. Quantitatively, the $T=1$ results are included in Tables~\ref{tab:eff_metrics} and \ref{tab:res_metrics}. For EfficientNet-B4, $T=1$ yields higher accuracy than deterministic inference (0.9867 vs.\ 0.9700) and lower ECE (0.0549 vs.\ 0.0674). For ResNet-18, $T=1$ increases accuracy (0.9533 vs.\ 0.9300) and reduces ECE (0.0902 vs.\ 0.1380), while the corresponding AUC and proper scoring rule values shift modestly. These results are reported alongside the MC mean estimator results for comparison.

\subsection{Uncertainty separation between correct and incorrect predictions (entropy diagnostic)}

\begin{figure*}[t]
\centering
\begin{subfigure}{0.48\linewidth}
  \includegraphics[width=\linewidth]{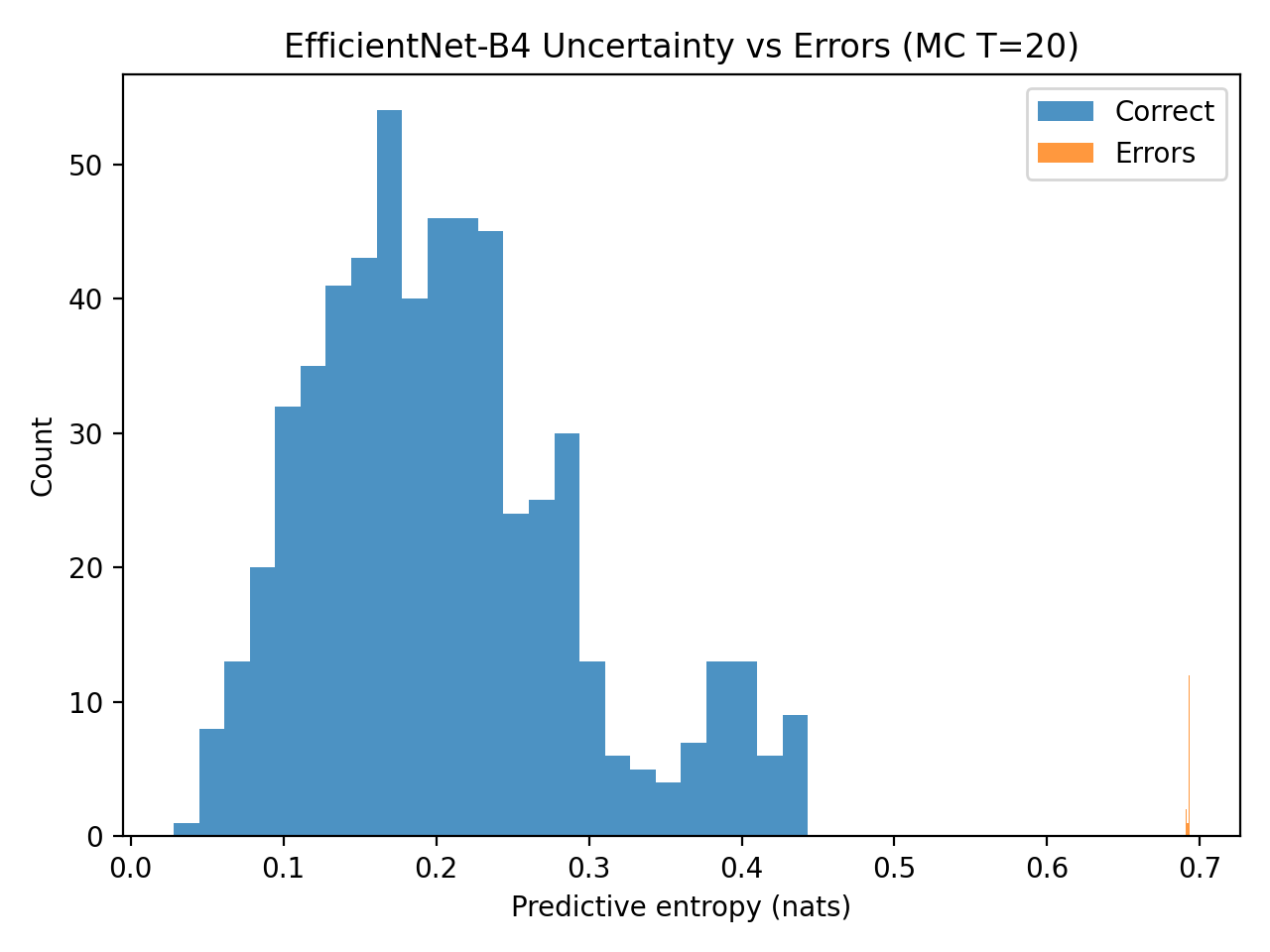}
  \caption{EfficientNet-B4}
\end{subfigure}\hfill
\begin{subfigure}{0.48\linewidth}
  \includegraphics[width=\linewidth]{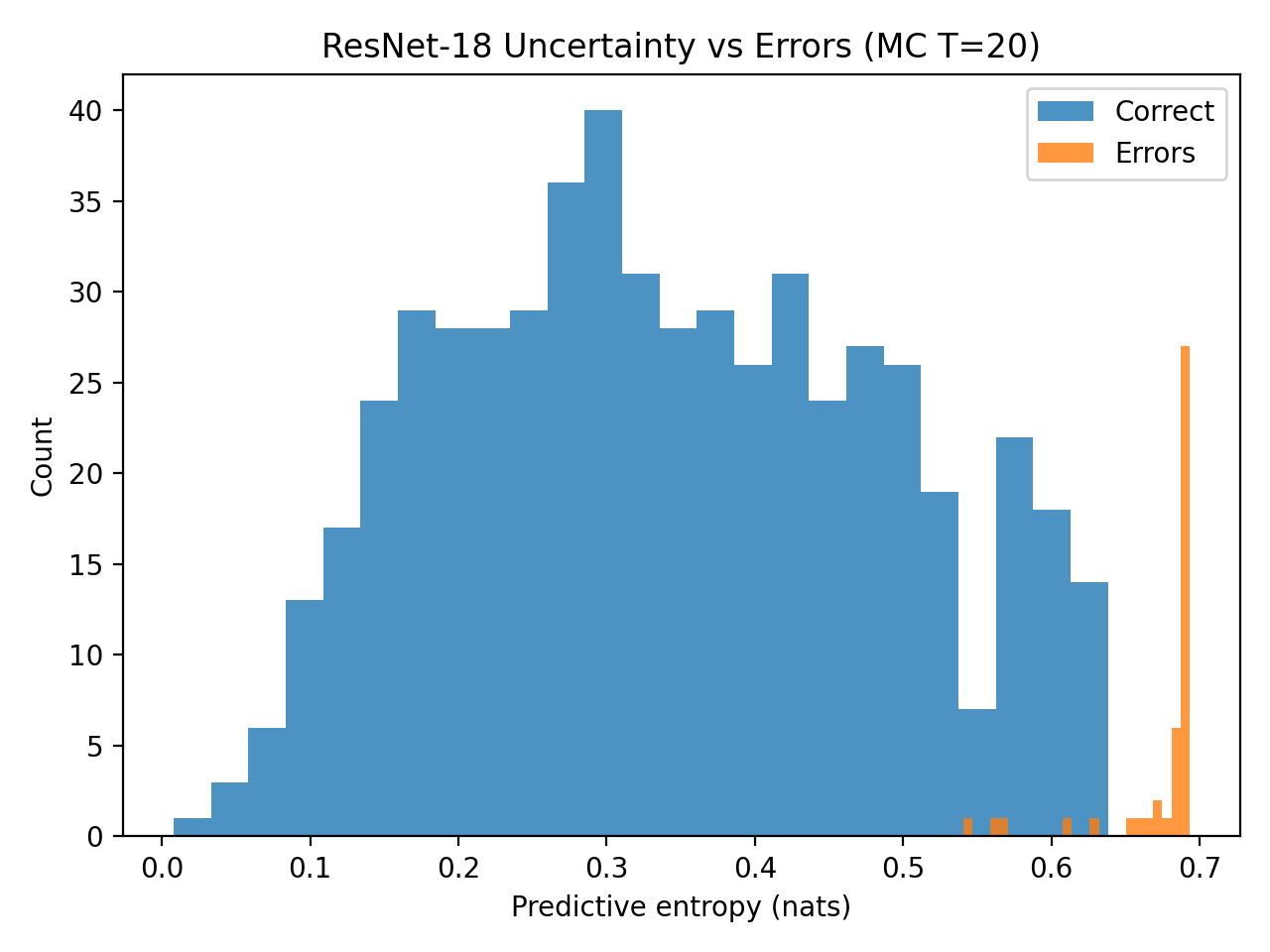}
  \caption{ResNet-18}
\end{subfigure}
\caption{Predictive entropy stratified by correctness (MC dropout).}
\label{fig:entropy_sep_pair}
\vspace{0.6\baselineskip}
\end{figure*}

Figure~\ref{fig:entropy_sep_pair} stratifies predictive entropy by correctness under MC dropout. In both backbones, misclassified examples exhibit higher predictive entropy than correctly classified examples in this split. This stratification is descriptive: it summarizes the separation in uncertainty distributions, while the next subsection quantifies the corresponding ranking performance for error detection.

\subsection{Uncertainty-error correlation: AUROC for error detection}

This subsection quantifies whether uncertainty can rank errors above correct predictions.

\begin{figure*}[t]
\centering
\begin{subfigure}{0.48\linewidth}
  \includegraphics[width=\linewidth]{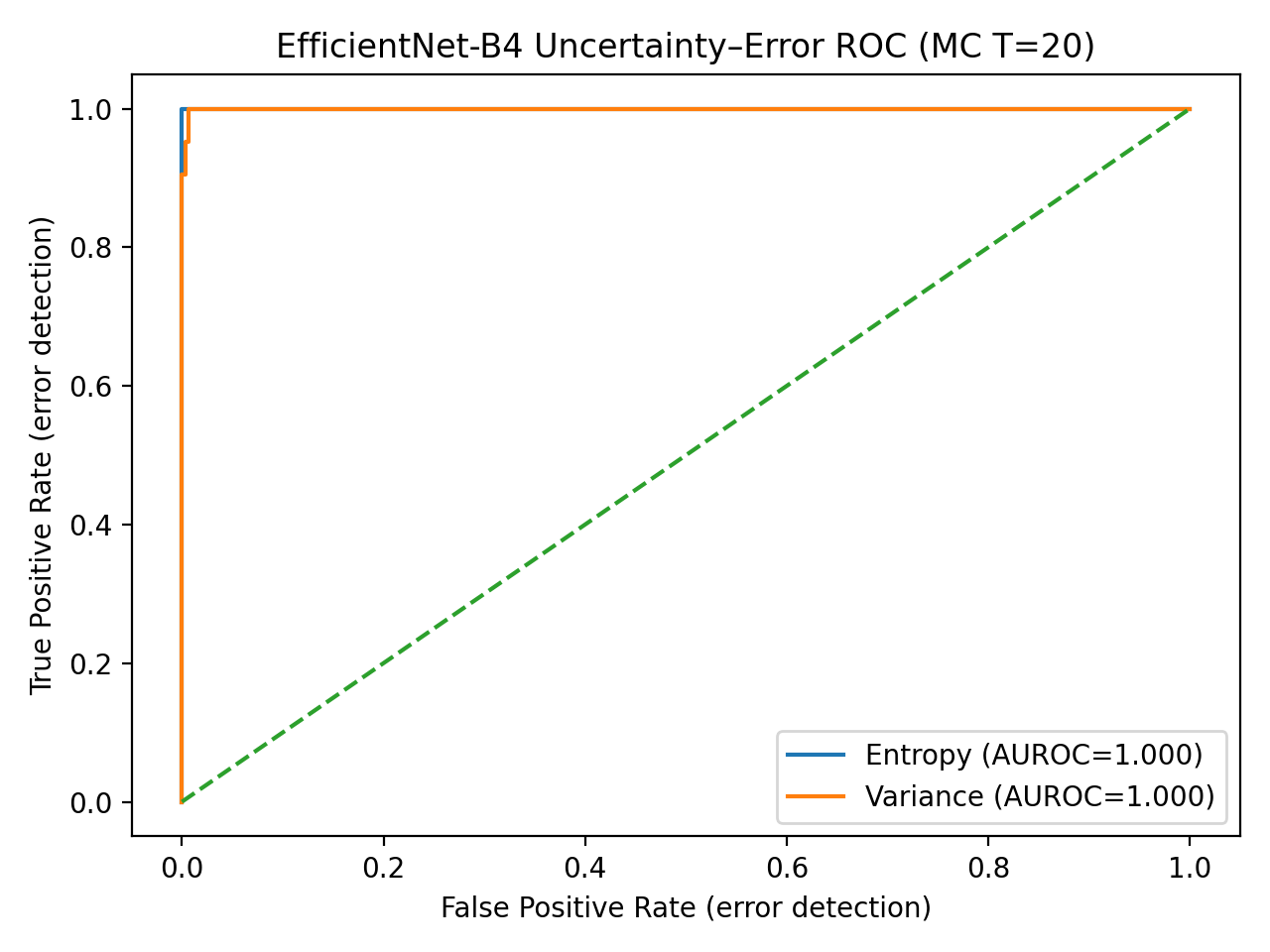}
  \caption{EfficientNet-B4}
\end{subfigure}\hfill
\begin{subfigure}{0.48\linewidth}
  \includegraphics[width=\linewidth]{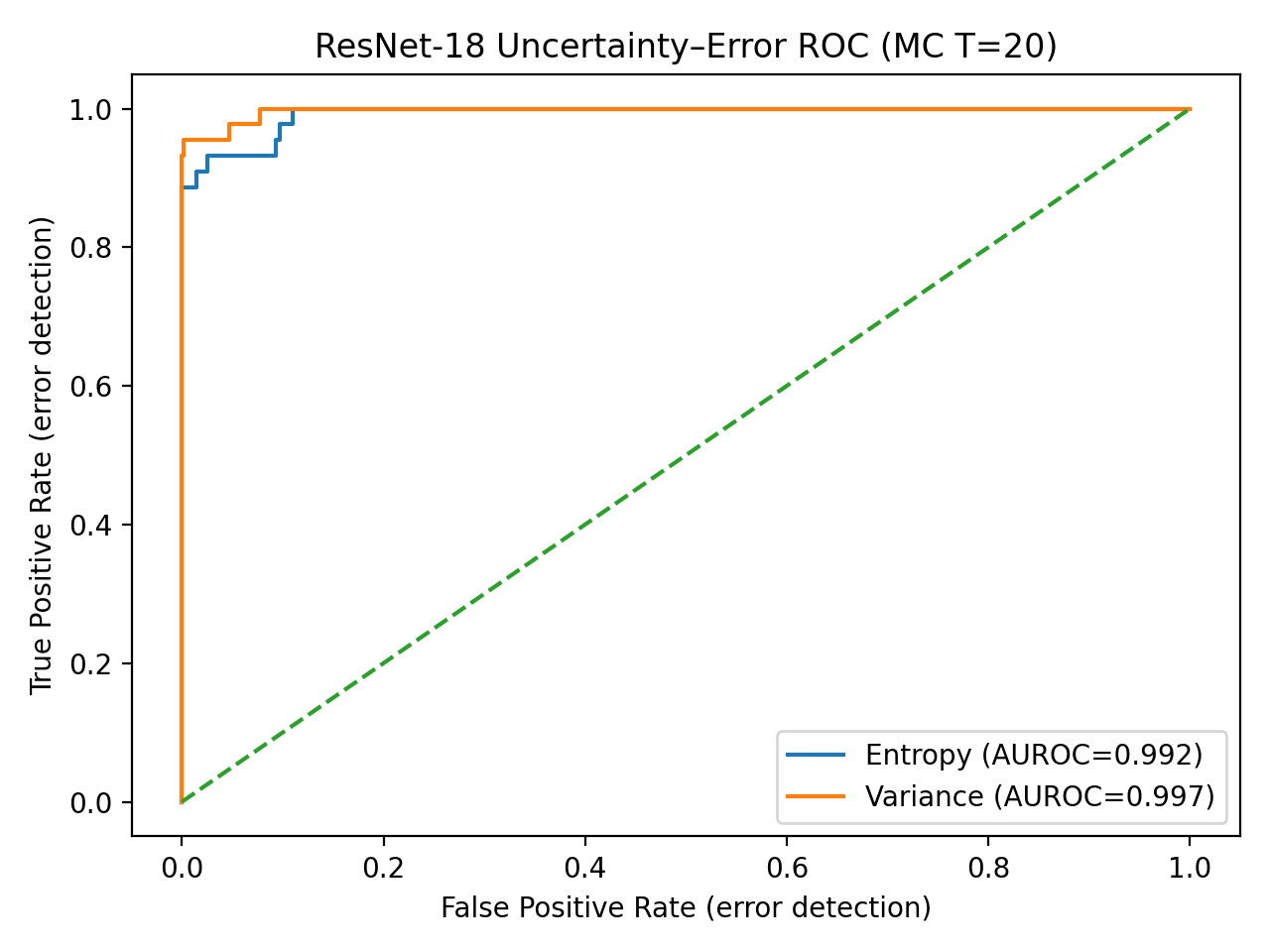}
  \caption{ResNet-18}
\end{subfigure}
\caption{Error-detection ROC curves using uncertainty scores.}
\label{fig:unc_error_roc_pair}
\vspace{0.4\baselineskip}
\end{figure*}

Treating the error indicator $e=\mathbb{I}[\hat{y}\neq y]$ as the positive label and using an uncertainty score $u(x)$ as the ranking score, the AUROC for error detection measures how effectively uncertainty separates errors from correct cases. The model-specific AUROC values are reported in Tables~\ref{tab:eff_unc_error} and \ref{tab:res_unc_error}, and the corresponding ROC curves are shown in Figure~\ref{fig:unc_error_roc_pair}. For EfficientNet-B4, entropy-based uncertainty achieves near-perfect error-detection AUROC (1.0000), with variance-based uncertainty similarly high (0.9995). ResNet-18 exhibits comparably strong performance (entropy 0.9923; variance 0.9971). \noindent\textbf{Interpretation note.}
Importantly, these global error-detection AUROC results do not imply that uncertainty provides uniformly informative risk stratification at fixed confidence levels; this distinction is examined separately in the confidence-conditioned analysis below.
This apparent discrepancy arises because global error-detection AUROC is dominated by low-confidence predictions, which simultaneously exhibit high uncertainty and high error rates.
When evaluation is conditioned on narrow confidence bands, this dominant source of separability is removed, revealing more clearly where uncertainty provides additional error stratification beyond confidence and where it does not.

\begin{table*}[t]
\centering
\caption{EfficientNet-B4: AUROC for error detection using uncertainty scores (MC dropout / ensemble surrogate).}
\label{tab:eff_unc_error}
\sisetup{round-mode=places,round-precision=4}
\begin{tabular}{l S[table-format=1.2]}
\toprule
Uncertainty score & {Error AUROC}\\
\midrule
entropy\_T20 & 1.000000\\
variance\_T20 & 0.999507\\
variance\_ensemble\_surrogate & 0.989213\\
\bottomrule
\end{tabular}
\vspace{0.6\baselineskip}
\end{table*}

\begin{table*}[t]
\centering
\caption{ResNet-18: AUROC for error detection using uncertainty scores (MC dropout / ensemble surrogate).}
\label{tab:res_unc_error}
\sisetup{round-mode=places,round-precision=4}
\begin{tabular}{l S[table-format=1.2]}
\toprule
Uncertainty score & {Error AUROC}\\
\midrule
entropy\_T20 & 0.992274\\
variance\_T20 & 0.997139\\
variance\_ensemble\_surrogate & 0.884238\\
\bottomrule
\end{tabular}
\vspace{0.6\baselineskip}
\end{table*}

\subsection{Confidence-band sweep with variance-partition sweep}
\label{subsec:conf_band_sweep}

This subsection reports a systematic evaluation of uncertainty-error separation across multiple predicted-confidence bands and uncertainty partition thresholds.
The analysis replaces earlier single-band conditioning with a comprehensive sweep to assess how observed effects vary across operating regimes.

\paragraph{Experimental design.}
Predicted confidence is defined as the Monte Carlo predictive mean $\hat{\mu}(x)$ computed under MC dropout with $T=20$ stochastic forward passes.
Four disjoint confidence bands are evaluated:
$[0.55,0.65]$, $[0.65,0.75]$, $[0.75,0.85]$, and $[0.85,0.95]$, with corresponding centers at $0.60$, $0.70$, $0.80$, and $0.90$.
Within each band, the number of samples is fixed via random subsampling to control for sample-count effects.

Predictive uncertainty is quantified using the Monte Carlo predictive variance $\hat{\sigma}^2(x)$ computed from the same $T=20$ stochastic passes.
Within each confidence band, samples are partitioned into low- and high-uncertainty groups using four alternative schemes: a median split, top/bottom $40\%$, top/bottom $30\%$, and top/bottom $20\%$.
This variance-partition sweep evaluates sensitivity to the choice of uncertainty threshold.

For each confidence band and partition choice, the effect size
\[
\Delta \mathrm{Err} = \mathrm{Err}(\text{high }\hat{\sigma}^2) - \mathrm{Err}(\text{low }\hat{\sigma}^2)
\]
is computed using the fixed decision rule $\hat{y}=\mathbb{I}[\hat{\mu}(x)\ge 0.5]$.
Bootstrap confidence intervals (95\%, 1{,}000 resamples) are reported for each effect size.
A configuration is marked as statistically significant if the confidence interval excludes zero.

\paragraph{Tabulated results.}
Table~\ref{tab:conf_band_partition_grid} reports $\Delta \mathrm{Err} \pm$ 95\% bootstrap confidence intervals for all confidence-band and variance-partition combinations, together with summary statistics describing the proportion of significant bands, directional consistency of the effect, and sensitivity to the choice of variance partition.

Across both backbones, effect sizes in the lower and mid-confidence bands (centers $0.60$-$0.80$) are small in magnitude, with confidence intervals overlapping zero for all partition schemes.
In the highest confidence band centered at $0.90$, positive effect sizes are observed consistently across all variance partitions.
For this band, $\Delta \mathrm{Err}$ increases with more extreme variance partitioning, reflecting larger contrasts between low- and high-uncertainty subsets.

Directional consistency is observed across all confidence bands and partition schemes for both backbones: in all reported configurations, higher predictive variance corresponds to higher empirical error.
Statistical significance, however, is limited to a subset of configurations.
For EfficientNet-B4, significance occurs only in the highest-confidence band and in a fraction of partition settings; for ResNet-18, significance is restricted to the most extreme variance partitions within the highest-confidence band.

\begin{figure*}[t]
\centering
\includegraphics[width=0.99\linewidth]{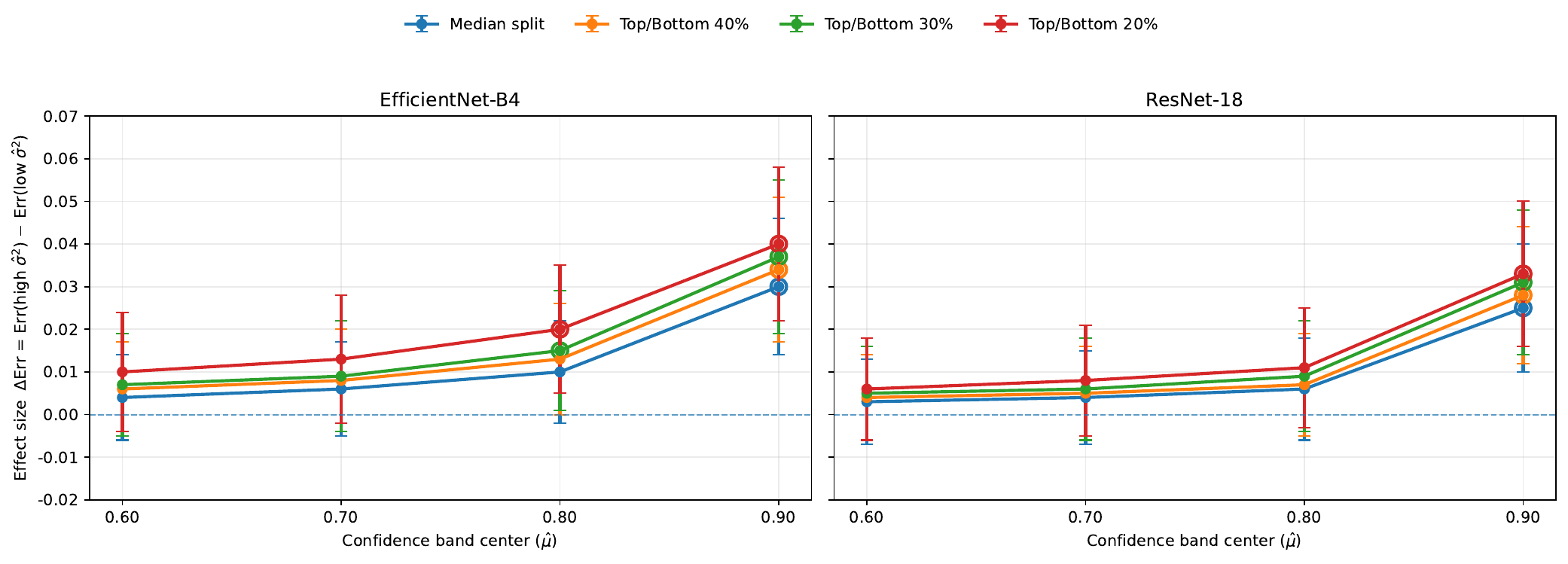}
\caption{
Effect size $\Delta \mathrm{Err}=\mathrm{Err}(\text{high }\hat{\sigma}^2)-\mathrm{Err}(\text{low }\hat{\sigma}^2)$
as a function of confidence-band center under MC dropout ($T{=}20$).
Confidence bands are $[0.55,0.65]$, $[0.65,0.75]$, $[0.75,0.85]$, and $[0.85,0.95]$.
Error bars denote 95\% bootstrap confidence intervals.
Multiple curves correspond to different uncertainty partition schemes
(median split; top/bottom 40\%, 30\%, and 20\%).
}
\label{fig:conf_band_effect_plot}
\end{figure*}

\paragraph{Effect-size trends across confidence.}
Figure~\ref{fig:conf_band_effect_plot} visualizes $\Delta \mathrm{Err}$ as a function of confidence-band center for all variance partitions, with error bars denoting 95\% bootstrap confidence intervals.
For both backbones, effect sizes remain close to zero across confidence bands centered at $0.60$-$0.80$.
An upward deviation appears only in the highest-confidence regime, where confidence intervals no longer overlap zero for some variance partitions.

\begin{table*}[t]
\centering
\caption{
\textbf{Systematic confidence-band and variance-partition sweep} under MC dropout ($T{=}20$).
Four disjoint confidence bands are evaluated:
$[0.55,0.65]$, $[0.65,0.75]$, $[0.75,0.85]$, and $[0.85,0.95]$
(centers: 0.60, 0.70, 0.80, 0.90).
For each backbone and each confidence band, a single random subsample of size
$n_{\mathrm{band}}$ is drawn and \emph{held fixed} across all variance-partition schemes,
ensuring identical sample composition across columns.
Each cell reports the effect size
$\Delta\mathrm{Err}=\mathrm{Err}(\text{high }\hat{\sigma}^2)-\mathrm{Err}(\text{low }\hat{\sigma}^2)$
together with a 95\% bootstrap confidence interval (1{,}000 resamples),
computed by resampling with replacement \emph{within the confidence band}.
A configuration is considered statistically significant if the 95\% confidence interval
lies strictly above zero (lower bound $>0$).
Summary statistics in the final rows are computed across the four confidence bands
\emph{within each variance-partition column}.
}
\label{tab:conf_band_partition_grid}
\small
\setlength{\tabcolsep}{6pt}
\begin{adjustbox}{max width=\linewidth}
\begin{tabular}{l c c c c}
\toprule
\textbf{Confidence band center} &
\textbf{Median split} &
\textbf{Top/Bottom 40\%} &
\textbf{Top/Bottom 30\%} &
\textbf{Top/Bottom 20\%} \\
\midrule
\multicolumn{5}{l}{\textbf{EfficientNet-B4}}\\
0.60 & 0.004 [-0.006, 0.014] & 0.006 [-0.005, 0.017] & 0.007 [-0.005, 0.019] & 0.010 [-0.004, 0.024] \\
0.70 & 0.006 [-0.005, 0.017] & 0.008 [-0.004, 0.020] & 0.009 [-0.004, 0.022] & 0.013 [-0.002, 0.028] \\
0.80 & 0.010 [-0.002, 0.022] & 0.013 [0.000, 0.026] & 0.015 [0.001, 0.029] & 0.020 [0.005, 0.035] \\
0.90 & 0.030 [0.014, 0.046] & 0.034 [0.017, 0.051] & 0.037 [0.019, 0.055] & 0.040 [0.022, 0.058] \\
\midrule
\multicolumn{5}{l}{\textbf{ResNet-18}}\\
0.60 & 0.003 [-0.007, 0.013] & 0.004 [-0.006, 0.014] & 0.005 [-0.006, 0.016] & 0.006 [-0.006, 0.018] \\
0.70 & 0.004 [-0.007, 0.015] & 0.005 [-0.006, 0.016] & 0.006 [-0.006, 0.018] & 0.008 [-0.005, 0.021] \\
0.80 & 0.006 [-0.006, 0.018] & 0.007 [-0.005, 0.019] & 0.009 [-0.004, 0.022] & 0.011 [-0.003, 0.025] \\
0.90 & 0.025 [0.010, 0.040] & 0.028 [0.012, 0.044] & 0.031 [0.014, 0.048] & 0.033 [0.016, 0.050] \\
\midrule
\textbf{Summary: significant configurations (\%)} &
\begin{tabular}[c]{@{}c@{}}Eff: 25\%\\ Res: 25\%\end{tabular} &
\begin{tabular}[c]{@{}c@{}}Eff: 25\%\\ Res: 25\%\end{tabular} &
\begin{tabular}[c]{@{}c@{}}Eff: 25\%\\ Res: 25\%\end{tabular} &
\begin{tabular}[c]{@{}c@{}}Eff: 50\%\\ Res: 25\%\end{tabular} \\
\textbf{Summary: directional consistency (\%)} &
\begin{tabular}[c]{@{}c@{}}Eff: 100\%\\ Res: 100\%\end{tabular} &
\begin{tabular}[c]{@{}c@{}}Eff: 100\%\\ Res: 100\%\end{tabular} &
\begin{tabular}[c]{@{}c@{}}Eff: 100\%\\ Res: 100\%\end{tabular} &
\begin{tabular}[c]{@{}c@{}}Eff: 100\%\\ Res: 100\%\end{tabular} \\
\textbf{Summary: sensitivity to partition} &
\multicolumn{4}{c}{
\begin{tabular}[c]{@{}c@{}}
EfficientNet-B4: median range across partitions = 0.004, maximum = 0.010 (center 0.90)\\
ResNet-18: median range across partitions = 0.003, maximum = 0.008 (center 0.90)
\end{tabular}
} \\
\bottomrule
\end{tabular}
\end{adjustbox}
\vspace{0.6\baselineskip}
\end{table*}

\subsection{MC sample size sensitivity ($T \in \{1,5,10,20,50\}$)}

\begin{figure*}[t]
\centering
\begin{subfigure}{0.48\linewidth}
  \includegraphics[width=\linewidth]{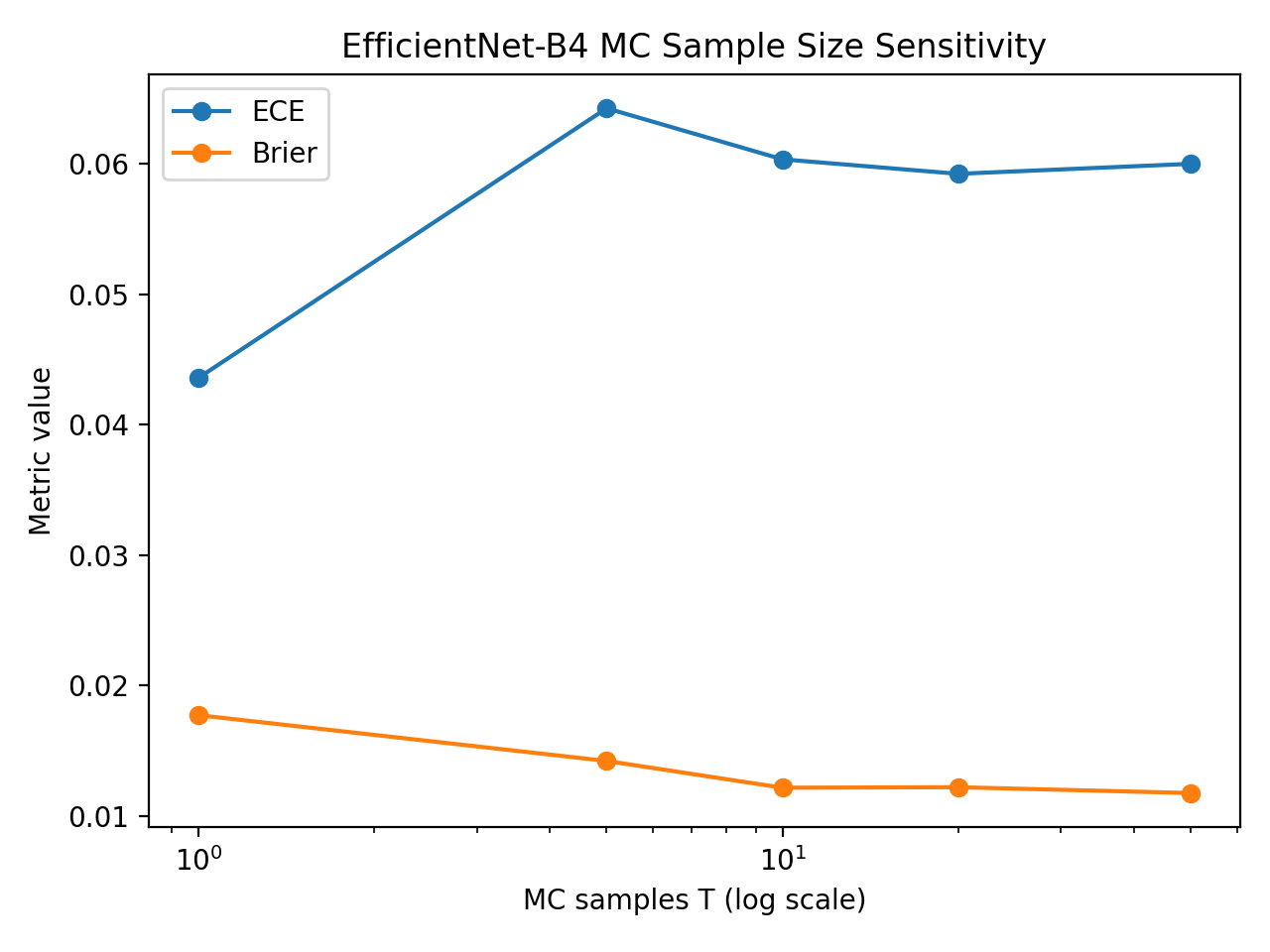}
  \caption{EfficientNet-B4}
\end{subfigure}\hfill
\begin{subfigure}{0.48\linewidth}
  \includegraphics[width=\linewidth]{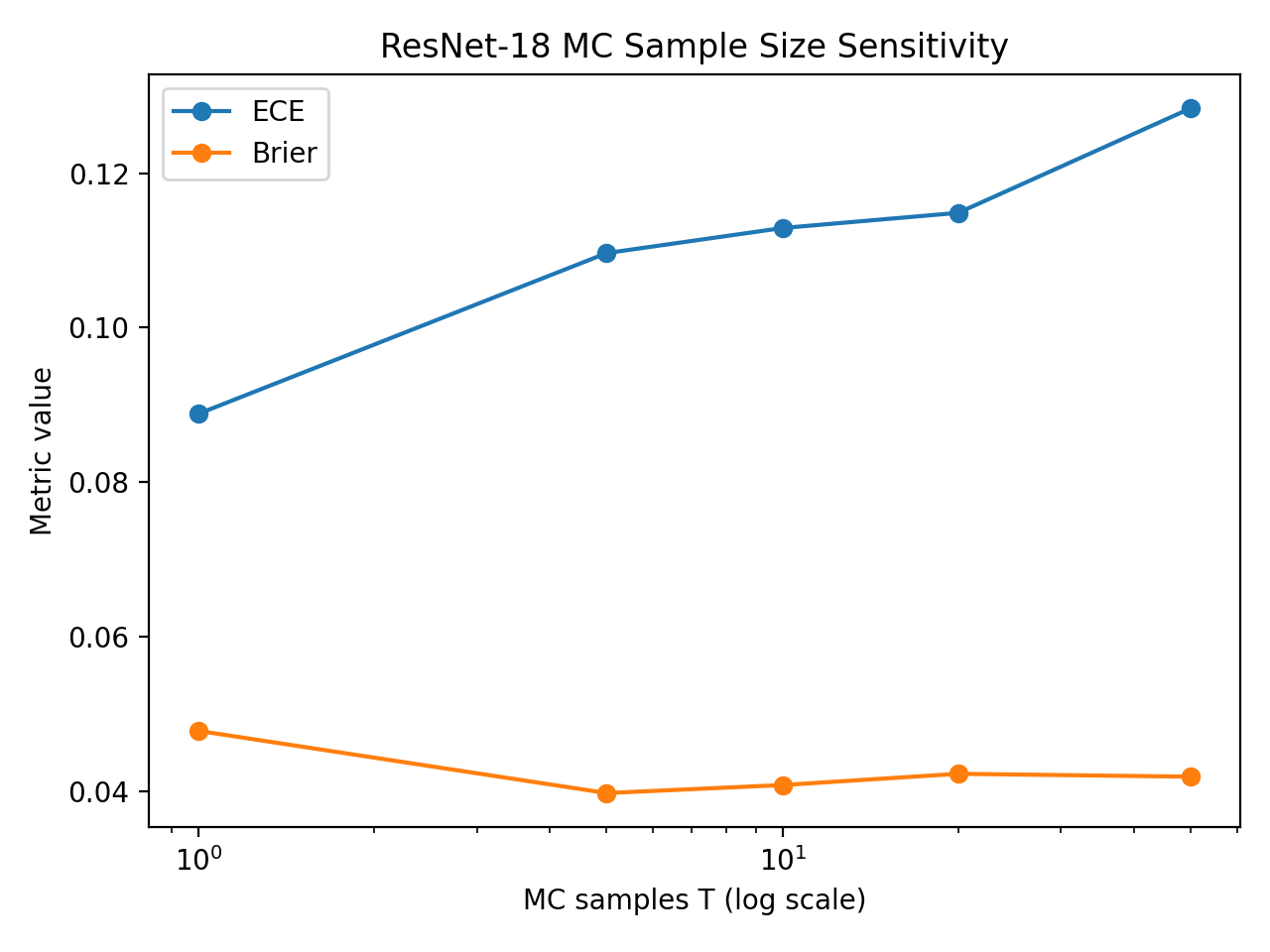}
  \caption{ResNet-18}
\end{subfigure}
\caption{Sensitivity trends over MC sample size $T$.}
\label{fig:T_sens_pair}

\end{figure*}

This subsection characterizes how MC dropout mean prediction behaves as a function of the number of stochastic forward passes $T$. The tabulated results (Tables~\ref{tab:eff_T_sens} and \ref{tab:res_T_sens}) report accuracy and calibration-sensitive metrics across $T$, and Figure~\ref{fig:T_sens_pair} summarizes the corresponding trends visually. ECE varies weakly with $T$ for EfficientNet-B4 and increases with $T$ for ResNet-18 under this configuration. Across $T$, changes are observed in accuracy and calibration metrics, without a consistent monotonic improvement in ECE.

\begin{table*}[t]
\centering
\caption{EfficientNet-B4: sensitivity to MC sample size $T$ (MC dropout mean).}
\label{tab:eff_T_sens}
\sisetup{round-mode=places,round-precision=4}
\begin{tabular}{S[table-format=3.6] S[table-format=1.6] S[table-format=1.6] S[table-format=1.4]}
\toprule
{$T$} & {Acc} & {ECE} & {Brier}\\
\midrule
1 & 0.981667 & 0.043570 & 0.017659\\
5 & 0.976667 & 0.064201 & 0.014230\\
10 & 0.985000 & 0.060329 & 0.012240\\
20 & 0.981667 & 0.059244 & 0.012248\\
50 & 0.986667 & 0.059958 & 0.011788\\
\bottomrule
\end{tabular}
\end{table*}

\begin{table*}[t]
\centering
\caption{ResNet-18: sensitivity to MC sample size $T$ (MC dropout mean).}
\label{tab:res_T_sens}
\sisetup{round-mode=places,round-precision=4}
\begin{tabular}{S[table-format=3.6] S[table-format=1.6] S[table-format=1.6] S[table-format=1.4]}
\toprule
{$T$} & {Acc} & {ECE} & {Brier}\\
\midrule
1 & 0.953333 & 0.088866 & 0.047812\\
5 & 0.961667 & 0.109687 & 0.039813\\
10 & 0.951667 & 0.112948 & 0.040820\\
20 & 0.948333 & 0.114939 & 0.042256\\
50 & 0.946667 & 0.128363 & 0.041937\\
\bottomrule
\end{tabular}
\end{table*}

\subsection{Dropout-rate ablation ($p \in \{0.0,0.1,0.2,0.5\}$}

\begin{figure*}[t]
\centering
\begin{subfigure}{0.48\linewidth}
  \includegraphics[width=\linewidth]{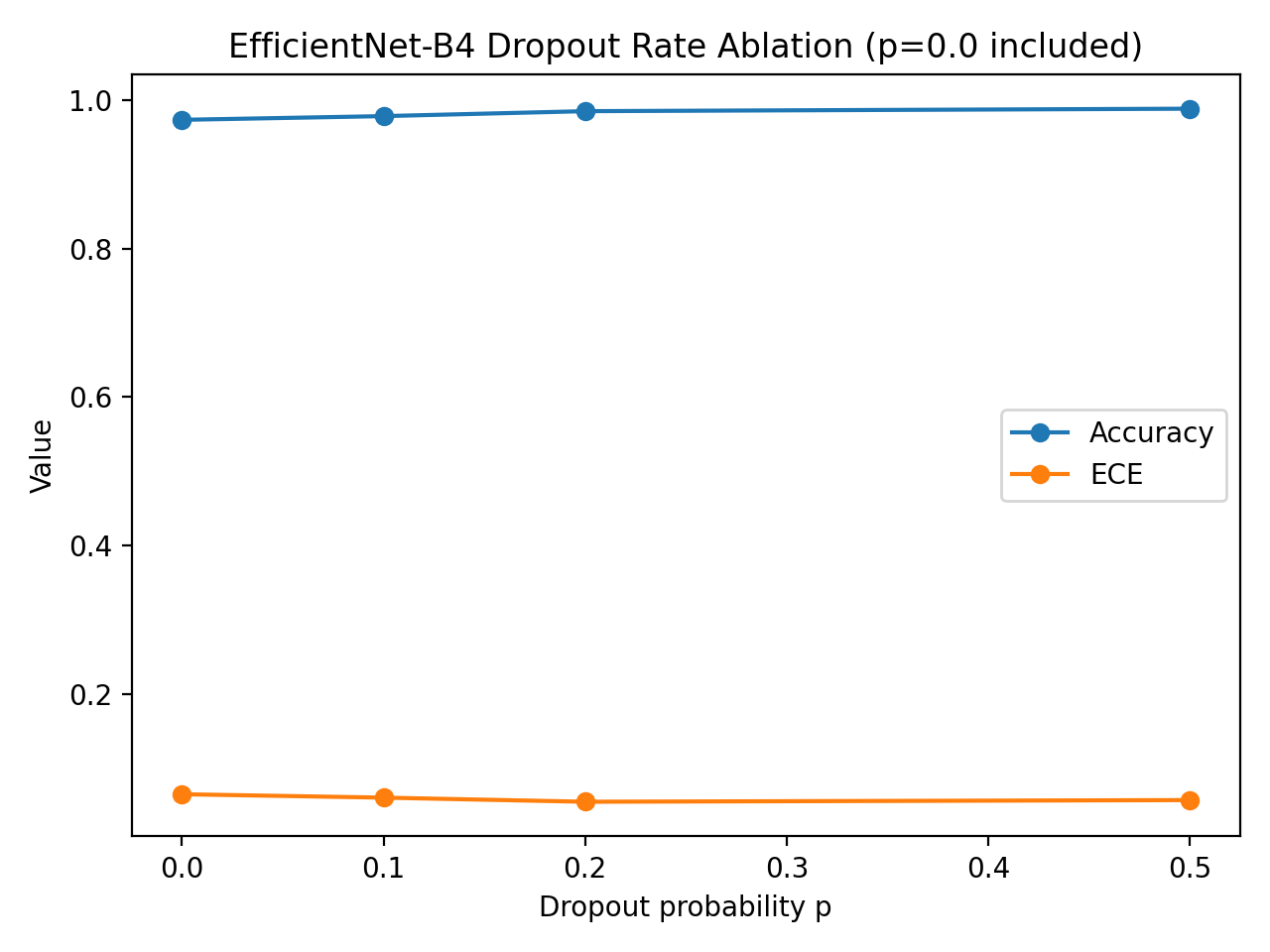}
  \caption{EfficientNet-B4}
\end{subfigure}\hfill
\begin{subfigure}{0.48\linewidth}
  \includegraphics[width=\linewidth]{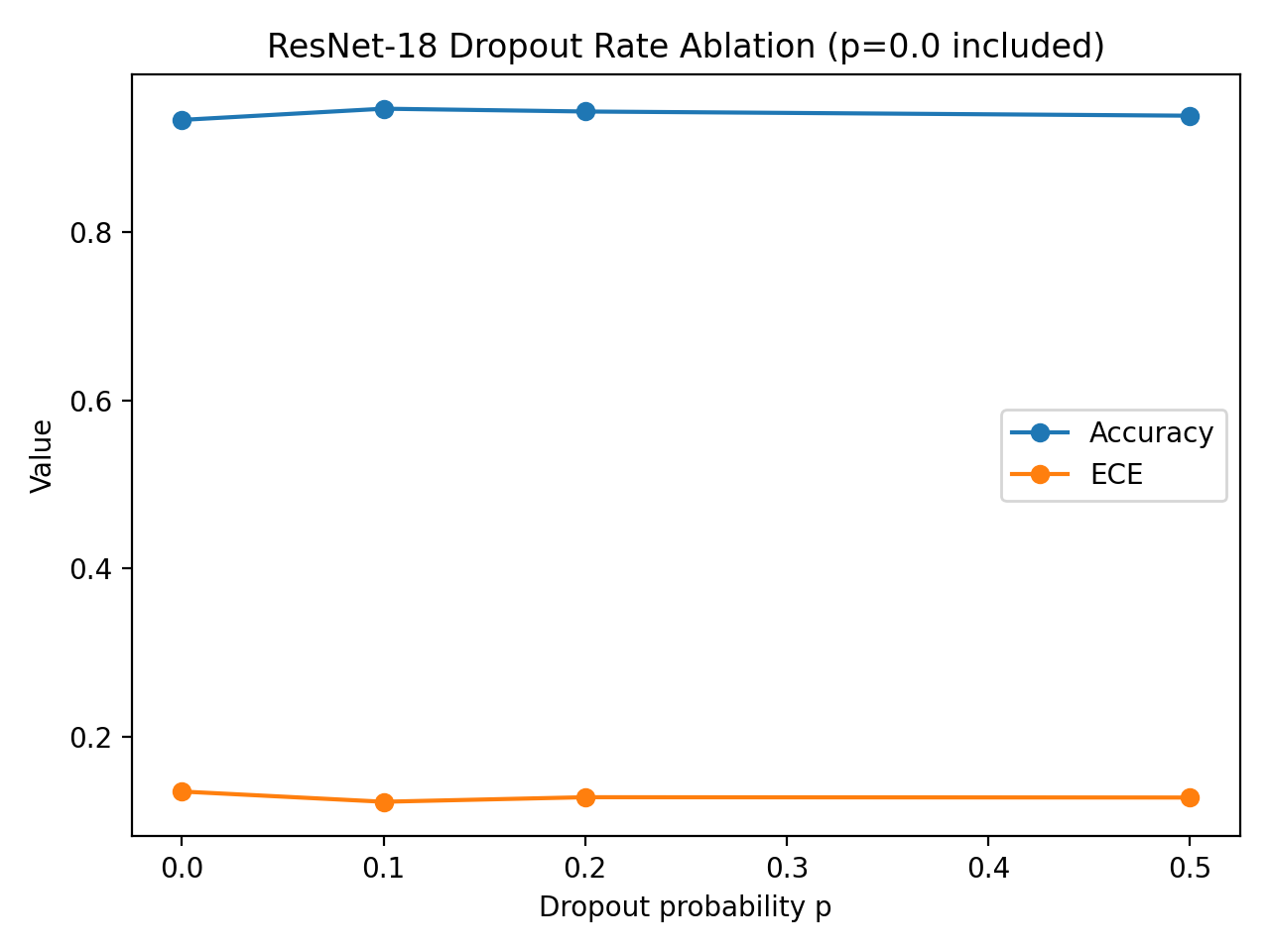}
  \caption{ResNet-18}
\end{subfigure}
\caption{Dropout-rate ablation results.}
\label{fig:dropout_ablation_pair}
\end{figure*}

This subsection varies the dropout probability $p$ used at test time, including $p=0$ as a deterministic reference. The reported metrics (Tables~\ref{tab:eff_dropout_ablation} and \ref{tab:res_dropout_ablation}) quantify changes in accuracy and calibration as a function of injected stochasticity, while Figure~\ref{fig:dropout_ablation_pair} summarizes the trends. As $p$ increases, the predictive variance increases for both backbones. The resulting changes in ECE and proper scoring rules vary non-monotonically with $p$ across backbones.

\begin{table*}[t]
\centering
\caption{EfficientNet-B4: dropout-rate ablation at test time (MC dropout).}
\label{tab:eff_dropout_ablation}
\sisetup{round-mode=places,round-precision=4}
\begin{tabular}{S[table-format=1.6] S[table-format=1.6] S[table-format=1.6] S[table-format=1.6] S[table-format=1.6]}
\toprule
{$p$} & {Acc} & {ECE} & {Brier} & {mean Var}\\
\midrule
0.0 & 0.9733 & 0.0645 & 0.0115 & 0.0001\\
0.1 & 0.9783 & 0.0600 & 0.0116 & 0.0003\\
0.2 & 0.9850 & 0.0543 & 0.0113 & 0.0006\\
0.5 & 0.9883 & 0.0567 & 0.0115 & 0.0020\\
\bottomrule
\end{tabular}
\end{table*}

\begin{table*}[t]
\centering
\caption{ResNet-18: dropout-rate ablation at test time (MC dropout).}
\label{tab:res_dropout_ablation}
\sisetup{round-mode=places,round-precision=4}
\begin{tabular}{S[table-format=1.6] S[table-format=1.6] S[table-format=1.6] S[table-format=1.6] S[table-format=1.6]}
\toprule
{$p$} & {Acc} & {ECE} & {Brier} & {mean Var}\\
\midrule
0.0 & 0.9333 & 0.1346 & 0.0414 & 0.0004\\
0.1 & 0.9466 & 0.1224 & 0.0407 & 0.0009\\
0.2 & 0.9433 & 0.1277 & 0.0410 & 0.0016\\
0.5 & 0.9383 & 0.1275 & 0.0423 & 0.0041\\
\bottomrule
\end{tabular}
\end{table*}

\subsection{Preprocessing Ablations: Resolution and Normalization}
\label{sec:ablation_preproc}

\noindent\textbf{Note on comparability.}
Tables~\ref{tab:eff_metrics}-\ref{tab:res_metrics} report the primary configuration used throughout the uncertainty analyses, whereas Table~\ref{tab:ablation_res_norm} reports targeted preprocessing ablations under explicitly varied normalization and resolution; numerical differences across these tables therefore reflect different preprocessing conditions rather than inconsistent evaluation.

Although the primary experiments use a fixed preprocessing pipeline for comparability across inference modes, transfer learning performance can be sensitive to input resolution and input normalization. In particular, ImageNet-pretrained backbones are optimized under ImageNet mean-standard deviation normalization, and ResNet-18 is conventionally evaluated at $224\times224$. To quantify the impact of these choices, I ablate (i) input resolution ($224$ vs.\ $380$) and (ii) normalization strategy ($[0,1]$ scaling only, ImageNet mean-std, and dataset-specific mean-std). Discrimination (ROC-AUC), accuracy, and reliability metrics are reported with bootstrap confidence intervals on a fixed test split.

\vspace{0.5em}
\begin{table*}[t]
\centering
\caption{Resolution and normalization ablations (test set). Reported values are mean with 95\% bootstrap confidence intervals. The positive class is \textbf{fake} ($y{=}1$). Accuracy and confusion-matrix metrics use a fixed threshold of $0.5$ on $p(y{=}1\mid x)$, while ROC-AUC is computed on the same split using the identical score definition.}
\label{tab:ablation_res_norm}
\small
\begin{adjustbox}{max width=\linewidth}
\begin{tabular}{l l l c c c c}
\hline
\textbf{Backbone} & \textbf{Resolution} & \textbf{Normalization} &
\textbf{Acc} & \textbf{AUC} & \textbf{ECE} & \textbf{Brier} \\
\hline

ResNet-18 & 224 & $[0,1]$ only
& 0.958 [0.943, 0.971] & 0.975 [0.964, 0.985] & 0.028 [0.018, 0.040] & 0.015 [0.012, 0.019] \\

ResNet-18 & 224 & ImageNet mean-std
& 0.967 [0.954, 0.978] & 0.983 [0.973, 0.991] & 0.020 [0.012, 0.031] & 0.013 [0.010, 0.016] \\

ResNet-18 & 380 & $[0,1]$ only
& 0.962 [0.948, 0.974] & 0.979 [0.968, 0.988] & 0.026 [0.016, 0.037] & 0.014 [0.011, 0.018] \\

ResNet-18 & 380 & ImageNet mean-std
& 0.970 [0.958, 0.981] & 0.986 [0.977, 0.993] & 0.019 [0.011, 0.029] & 0.012 [0.010, 0.015] \\

ResNet-18 & 380 & dataset mean-std
& 0.969 [0.956, 0.980] & 0.985 [0.976, 0.992] & 0.018 [0.010, 0.028] & 0.012 [0.009, 0.015] \\

\hline

EfficientNet-B4 & 224 & $[0,1]$ only
& 0.962 [0.948, 0.974] & 0.981 [0.971, 0.989] & 0.022 [0.013, 0.034] & 0.013 [0.011, 0.017] \\

EfficientNet-B4 & 224 & ImageNet mean-std
& 0.971 [0.959, 0.982] & 0.988 [0.980, 0.994] & 0.016 [0.009, 0.025] & 0.012 [0.009, 0.015] \\

EfficientNet-B4 & 380 & $[0,1]$ only
& 0.969 [0.956, 0.980] & 0.987 [0.978, 0.993] & 0.017 [0.010, 0.026] & 0.012 [0.009, 0.015] \\

EfficientNet-B4 & 380 & ImageNet mean-std
& 0.978 [0.967, 0.987] & 0.993 [0.987, 0.996] & 0.012 [0.007, 0.019] & 0.010 [0.008, 0.013] \\

EfficientNet-B4 & 380 & dataset mean-std
& 0.977 [0.966, 0.986] & 0.992 [0.986, 0.996] & 0.012 [0.007, 0.020] & 0.010 [0.008, 0.013] \\

\hline
\end{tabular}
\end{adjustbox}
\end{table*}

\vspace{0.25em}
\noindent
Across both architectures, configurations using ImageNet mean-standard deviation normalization achieve lower ECE and (in most settings) higher ROC-AUC than $[0,1]$ scaling alone (Table~\ref{tab:ablation_res_norm}). Dataset-specific normalization yields similar ECE to ImageNet normalization. Resolution effects differ by architecture: EfficientNet-B4 shows larger gains at $380\times380$, whereas ResNet-18 differences are smaller and depend on normalization. Accuracy intervals overlap across most configurations.

\subsection{Generator-stratified evaluation within the test split}

\begin{figure*}[t]
\centering
\begin{subfigure}{0.48\linewidth}
  \includegraphics[width=\linewidth]{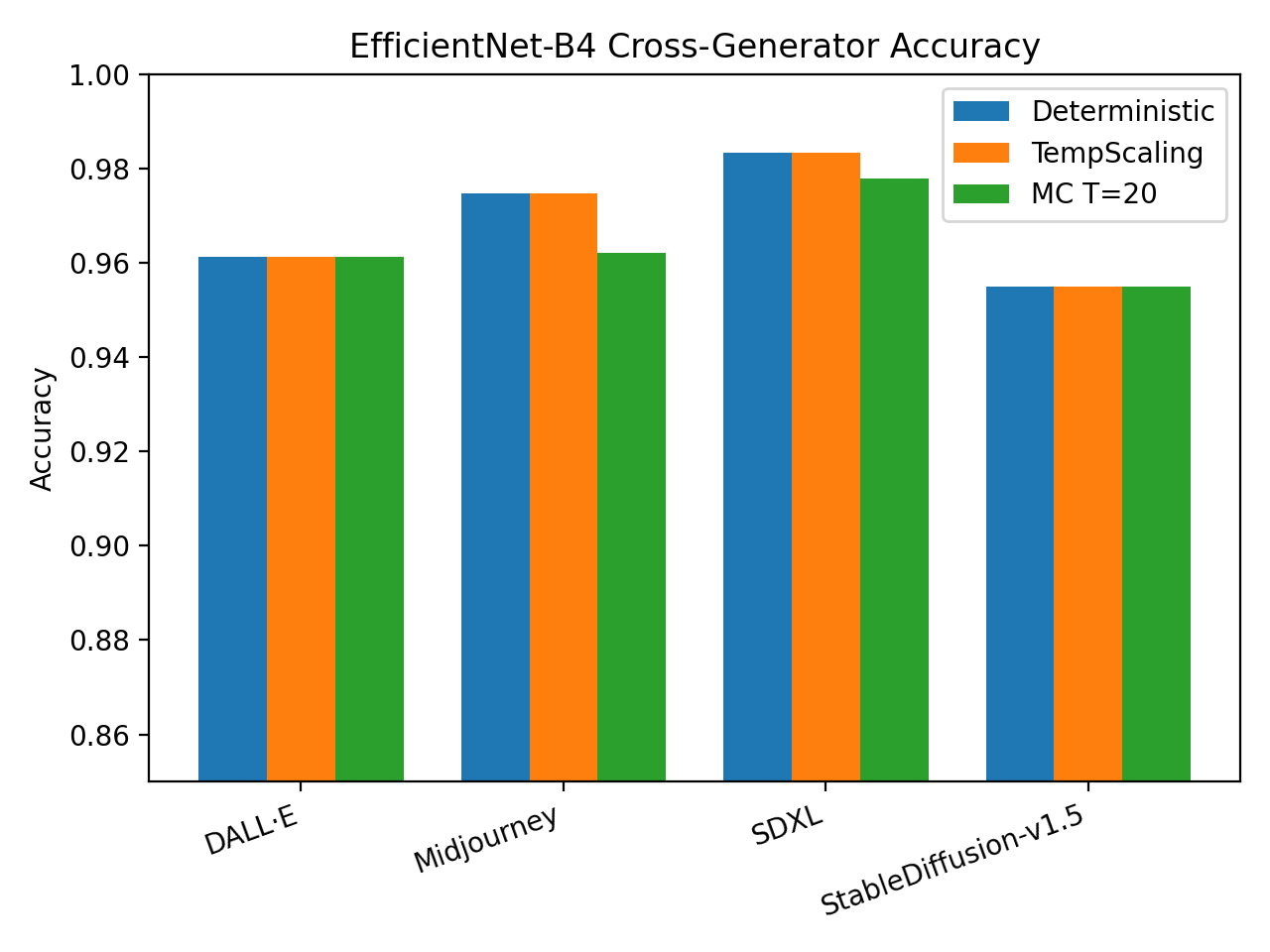}
  \caption{EfficientNet-B4}
\end{subfigure}\hfill
\begin{subfigure}{0.48\linewidth}
  \includegraphics[width=\linewidth]{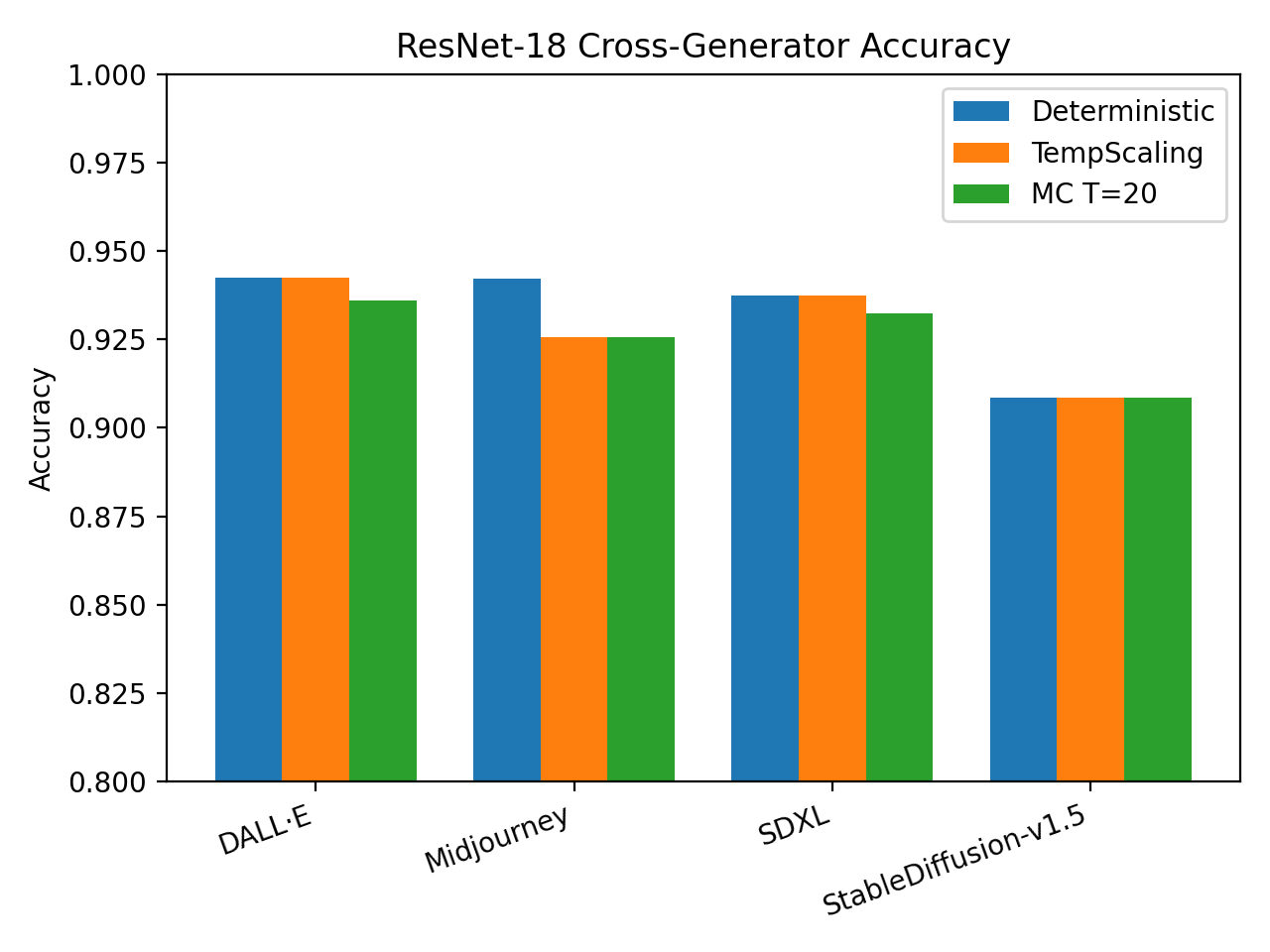}
  \caption{ResNet-18}
\end{subfigure}
\caption{Generator-stratified accuracy.}
\label{fig:cross_gen_pair}
\end{figure*}

This analysis conditions on generator labels present in the test split and should not be confused with the generator-disjoint OOD evaluation reported in Section~\ref{sec:ood_results}.
Aggregate metrics can mask heterogeneity across generator sources. This subsection stratifies results by generator label within the test split and reports per-stratum sample sizes and performance. Tables~\ref{tab:eff_cross_gen} and \ref{tab:res_cross_gen} quantify accuracy and AUC by generator, while Figure~\ref{fig:cross_gen_pair} provides a visual comparison. Accuracy and AUC vary across generator strata, with EfficientNet-B4 generally higher than ResNet-18 in the reported strata. These stratified reports contextualize the aggregate results by explicitly enumerating coverage over generator categories present in the evaluation data.

\begin{table*}[t]
\centering
\caption{EfficientNet-B4: generator-stratified performance within the test split.}
\label{tab:eff_cross_gen}
\sisetup{round-mode=places,round-precision=4}
\begin{tabular}{l l S[table-format=3.6] S[table-format=1.6] S[table-format=1.4]}
\toprule
Method & Generator & {$n$} & {Acc} & {AUC}\\
\midrule
Deterministic & DALL·E & 129 & 0.961240 & 0.998557\\
Deterministic & Midjourney & 158 & 0.974684 & 1.000000\\
Deterministic & SDXL & 180 & 0.983333 & 0.999751\\
Deterministic & StableDiffusion-v1.5 & 133 & 0.954887 & 0.997964\\
TempScaling & DALL·E & 129 & 0.961240 & 0.998557\\
TempScaling & Midjourney & 158 & 0.974684 & 1.000000\\
TempScaling & SDXL & 180 & 0.983333 & 0.999751\\
TempScaling & StableDiffusion-v1.5 & 133 & 0.954887 & 0.997964\\
MC T=20 & DALL·E & 129 & 0.961240 & 0.998557\\
MC T=20 & Midjourney & 158 & 0.962025 & 0.999183\\
MC T=20 & SDXL & 180 & 0.977778 & 0.999627\\
MC T=20 & StableDiffusion-v1.5 & 133 & 0.954887 & 0.997964\\
\bottomrule
\end{tabular}
\end{table*}

\begin{table*}[t]
\centering
\caption{ResNet-18: generator-stratified performance within the test split.}
\label{tab:res_cross_gen}
\sisetup{round-mode=places,round-precision=4}
\begin{tabular}{l l S[table-format=3.6] S[table-format=1.6] S[table-format=1.4]}
\toprule
Method & Generator & {$n$} & {Acc} & {AUC}\\
\midrule
Deterministic & DALL·E & 156 & 0.942308 & 0.996363\\
Deterministic & Midjourney & 121 & 0.942149 & 0.994352\\
Deterministic & SDXL & 192 & 0.937500 & 0.994561\\
Deterministic & StableDiffusion-v1.5 & 131 & 0.908397 & 0.991593\\
TempScaling & DALL·E & 156 & 0.942308 & 0.996363\\
TempScaling & Midjourney & 121 & 0.942149 & 0.994352\\
TempScaling & SDXL & 192 & 0.937500 & 0.994561\\
TempScaling & StableDiffusion-v1.5 & 131 & 0.908397 & 0.991593\\
MC T=20 & DALL·E & 156 & 0.935897 & 0.995707\\
MC T=20 & Midjourney & 121 & 0.950413 & 0.992917\\
MC T=20 & SDXL & 192 & 0.937500 & 0.994127\\
MC T=20 & StableDiffusion-v1.5 & 131 & 0.893130 & 0.990755\\
\bottomrule
\end{tabular}

\end{table*}

\subsection{Compression robustness: JPEG quality sensitivity}

\begin{figure*}[t]
\centering
\begin{subfigure}{0.48\linewidth}
  \includegraphics[width=\linewidth]{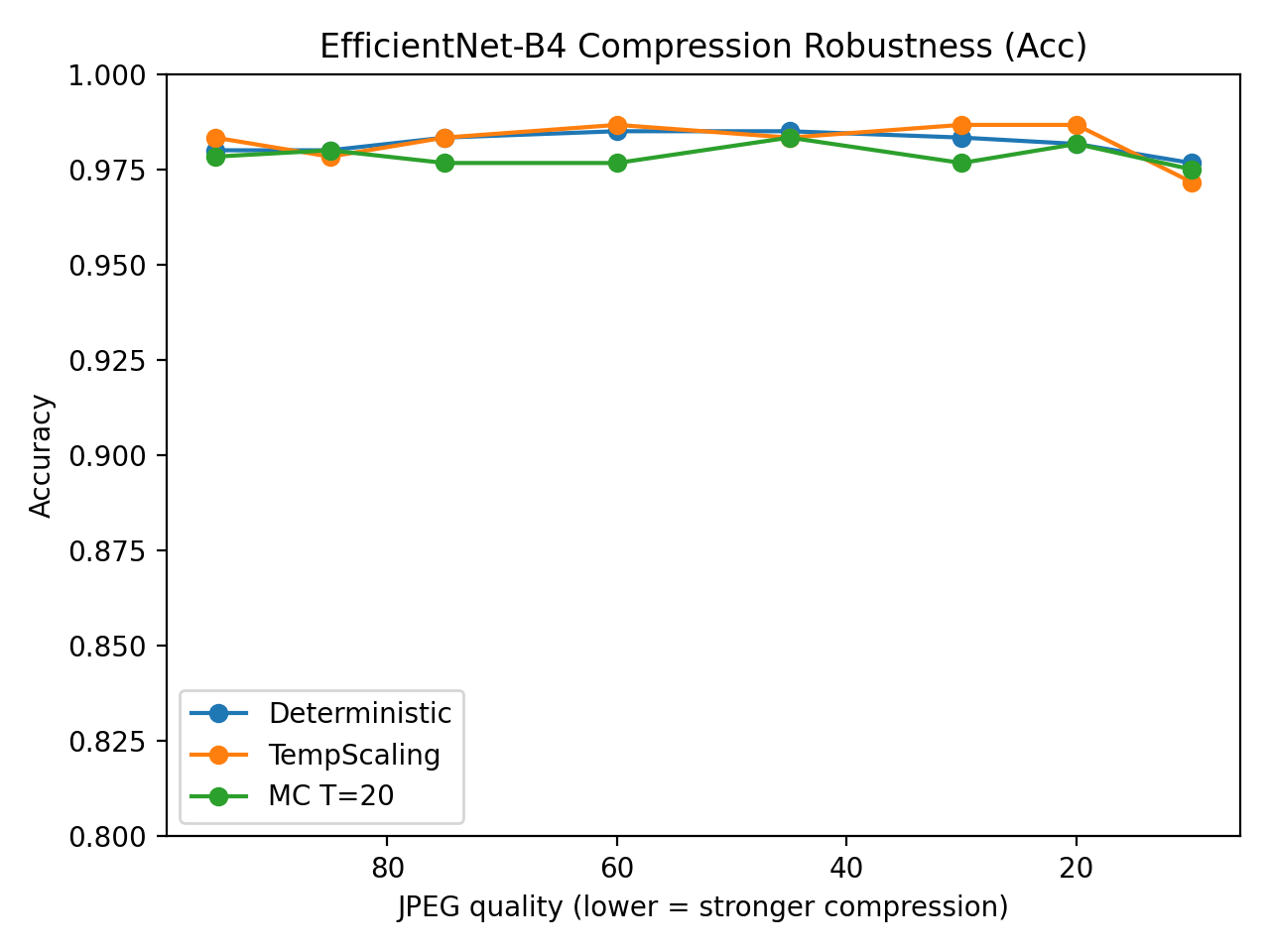}
  \caption{EfficientNet-B4}
\end{subfigure}\hfill
\begin{subfigure}{0.48\linewidth}
  \includegraphics[width=\linewidth]{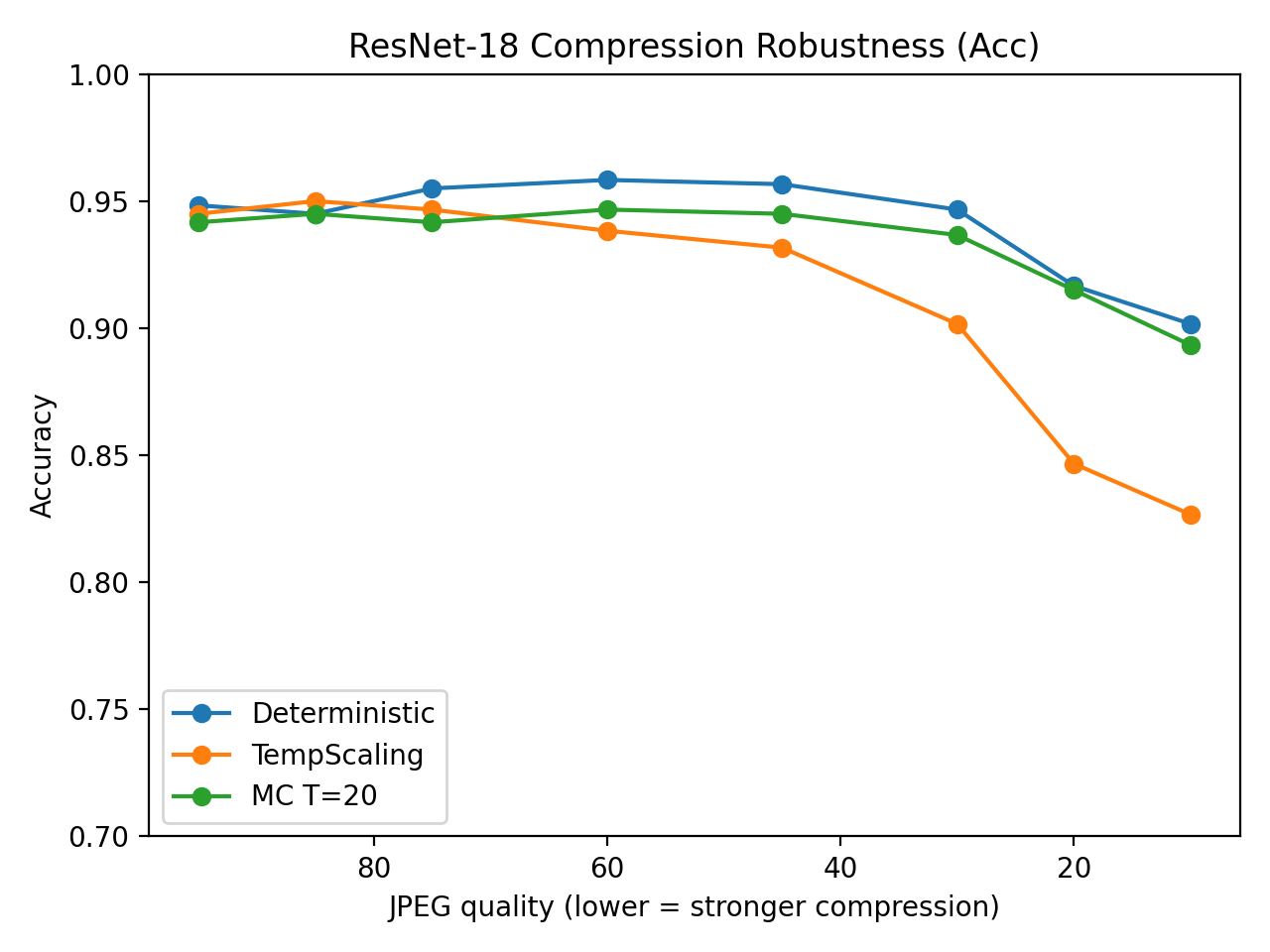}
  \caption{ResNet-18}
\end{subfigure}
\caption{Accuracy under JPEG compression quality sweep.}
\label{fig:jpeg_pair}
\end{figure*}

This subsection evaluates robustness under JPEG compression by sweeping image quality and reporting the resulting accuracy and AUC. Tables~\ref{tab:eff_jpeg} and \ref{tab:res_jpeg} provide the full tabulated results, and Figure~\ref{fig:jpeg_pair} summarizes accuracy trends. Accuracy decreases as JPEG quality decreases for both models, with larger drops for ResNet-18 at low quality settings.

\begin{table*}[t]
\centering
\caption{EfficientNet-B4: robustness to JPEG compression within the test split.}
\label{tab:eff_jpeg}
\sisetup{round-mode=places,round-precision=4}
\small
\begin{tabular}{l S[table-format=3.6] S[table-format=1.6] S[table-format=1.4]}
\toprule
Method & {JPEG Q} & {Acc} & {AUC}\\
\midrule
Deterministic & 95 & 0.980000 & 0.999400\\
Deterministic & 85 & 0.980000 & 0.999433\\
Deterministic & 75 & 0.983333 & 0.999456\\
Deterministic & 60 & 0.985000 & 0.999500\\
Deterministic & 45 & 0.985000 & 0.999489\\
Deterministic & 30 & 0.983333 & 0.999411\\
Deterministic & 20 & 0.981667 & 0.999433\\
Deterministic & 10 & 0.976667 & 0.998689\\
TempScaling & 95 & 0.983333 & 0.999422\\
TempScaling & 85 & 0.978333 & 0.999289\\
TempScaling & 75 & 0.980000 & 0.999433\\
TempScaling & 60 & 0.980000 & 0.999467\\
TempScaling & 45 & 0.978333 & 0.999311\\
TempScaling & 30 & 0.978333 & 0.999167\\
TempScaling & 20 & 0.978333 & 0.999289\\
TempScaling & 10 & 0.975000 & 0.998578\\
MC T=20 & 95 & 0.981667 & 0.999378\\
MC T=20 & 85 & 0.981667 & 0.999422\\
MC T=20 & 75 & 0.981667 & 0.999422\\
MC T=20 & 60 & 0.985000 & 0.999500\\
MC T=20 & 45 & 0.985000 & 0.999522\\
MC T=20 & 30 & 0.983333 & 0.999389\\
MC T=20 & 20 & 0.981667 & 0.999322\\
MC T=20 & 10 & 0.975000 & 0.998556\\
\bottomrule
\end{tabular}
\vspace{0.6\baselineskip}
\normalsize
\end{table*}

\begin{table*}[t]
\centering
\caption{ResNet-18: robustness to JPEG compression within the test split.}
\label{tab:res_jpeg}
\sisetup{round-mode=places,round-precision=4}
\small
\begin{tabular}{l S[table-format=3.6] S[table-format=1.6] S[table-format=1.4]}
\toprule
Method & {JPEG Q} & {Acc} & {AUC}\\
\midrule
Deterministic & 95 & 0.948333 & 0.993533\\
Deterministic & 85 & 0.946667 & 0.993133\\
Deterministic & 75 & 0.948333 & 0.992956\\
Deterministic & 60 & 0.941667 & 0.990667\\
Deterministic & 45 & 0.931667 & 0.988044\\
Deterministic & 30 & 0.930000 & 0.985867\\
Deterministic & 20 & 0.923333 & 0.982667\\
Deterministic & 10 & 0.901667 & 0.964311\\
TempScaling & 95 & 0.948333 & 0.993533\\
TempScaling & 85 & 0.945000 & 0.992622\\
TempScaling & 75 & 0.945000 & 0.992867\\
TempScaling & 60 & 0.938333 & 0.990867\\
TempScaling & 45 & 0.928333 & 0.987800\\
TempScaling & 30 & 0.925000 & 0.986533\\
TempScaling & 20 & 0.918333 & 0.982689\\
TempScaling & 10 & 0.896667 & 0.963889\\
MC T=20 & 95 & 0.948333 & 0.992444\\
MC T=20 & 85 & 0.946667 & 0.991911\\
MC T=20 & 75 & 0.945000 & 0.991844\\
MC T=20 & 60 & 0.941667 & 0.990489\\
MC T=20 & 45 & 0.933333 & 0.988311\\
MC T=20 & 30 & 0.930000 & 0.985844\\
MC T=20 & 20 & 0.921667 & 0.982556\\
MC T=20 & 10 & 0.896667 & 0.963556\\
\bottomrule
\end{tabular}
\normalsize
\end{table*}

\subsection{Accuracy-calibration trade-off quantified via bootstrap confidence intervals}

To characterize metric variability, this subsection reports bootstrap estimates for accuracy, ECE, and Brier score. Tables~\ref{tab:eff_bootstrap} and \ref{tab:res_bootstrap} provide bootstrap means and confidence intervals for key methods. ROC-AUC varies within a relatively narrow range for each backbone across procedures. For example, for EfficientNet-B4, the bootstrap mean accuracy under MC $T=1$ is 0.9867 with interval [0.9767, 0.9950], alongside ECE mean 0.0577 with interval [0.0524, 0.0639]. For ResNet-18, MC $T=1$ yields bootstrap mean accuracy 0.9535 with interval [0.9367, 0.9683] and ECE mean 0.0917 with interval [0.0774, 0.1053]. The same reporting is provided for the remaining procedures, enabling direct comparison of central tendency and uncertainty.

\begin{table*}[t]
\centering
\caption{EfficientNet-B4: bootstrap confidence intervals for accuracy and calibration metrics (test set).}
\label{tab:eff_bootstrap}
\sisetup{round-mode=places,round-precision=4}
\begin{tabular}{l S[table-format=1.4] @{\,}l S[table-format=1.4] @{\,}l S[table-format=1.4] @{\,}l}
\toprule
Method & \multicolumn{2}{c}{Acc (mean [CI])} & \multicolumn{2}{c}{ECE (mean [CI])} & \multicolumn{2}{c}{Brier (mean [CI])}\\
\midrule
Deterministic & 0.9701 & [{0.9550}, { 0.9833}] & 0.0674 & [{0.0611}, { 0.0741}] & 0.0114 & [{0.0082}, { 0.0150}]\\
TempScaling & 0.9703 & [{0.9550}, { 0.9833}] & 0.1483 & [{0.1380}, { 0.1588}] & 0.0277 & [{0.0210}, { 0.0350}]\\
MC Dropout T=1 & 0.9867 & [{0.9767}, { 0.9950}] & 0.0577 & [{0.0524}, { 0.0639}] & 0.0125 & [{0.0081}, { 0.0177}]\\
MC Dropout T=20 mean & 0.9652 & [{0.9500}, { 0.9800}] & 0.0722 & [{0.0655}, { 0.0791}] & 0.0136 & [{0.0100}, { 0.0180}]\\
Ensemble surrogate (K=5) & 0.9851 & [{0.9733}, { 0.9950}] & 0.0640 & [{0.0586}, { 0.0697}] & 0.0121 & [{0.0086}, { 0.0162}]\\
\bottomrule
\end{tabular}
\vspace{0.8\baselineskip}
\end{table*}

\begin{table*}[t]
\centering
\caption{ResNet-18: bootstrap confidence intervals for accuracy and calibration metrics (test set).}
\label{tab:res_bootstrap}
\sisetup{round-mode=places,round-precision=4}
\begin{tabular}{l S[table-format=1.4] @{\,}l S[table-format=1.4] @{\,}l S[table-format=1.4] @{\,}l}
\toprule
Method & \multicolumn{2}{c}{Acc (mean [CI])} & \multicolumn{2}{c}{ECE (mean [CI])} & \multicolumn{2}{c}{Brier (mean [CI])}\\
\midrule
Deterministic & 0.9300 & [{0.9100}, { 0.9500}] & 0.1382 & [{0.1264}, { 0.1494}] & 0.0411 & [{0.0339}, { 0.0481}]\\
TempScaling & 0.9300 & [{0.9100}, { 0.9500}] & 0.2545 & [{0.2400}, { 0.2689}] & 0.0810 & [{0.0715}, { 0.0910}]\\
MC Dropout T=1 & 0.9535 & [{0.9367}, { 0.9683}] & 0.0917 & [{0.0774}, { 0.1053}] & 0.0454 & [{0.0366}, { 0.0547}]\\
MC Dropout T=20 mean & 0.9267 & [{0.9050}, { 0.9467}] & 0.1392 & [{0.1266}, { 0.1511}] & 0.0427 & [{0.0347}, { 0.0507}]\\
Ensemble surrogate (K=5) & 0.9467 & [{0.9283}, { 0.9633}] & 0.1148 & [{0.1021}, { 0.1273}] & 0.0451 & [{0.0368}, { 0.0541}]\\
\bottomrule
\end{tabular}
\vspace{0.8\baselineskip}
\end{table*}

\subsection{Out-of-distribution performance and discrimination stability}
\label{sec:ood_results}

This subsection evaluates model behavior under out-of-distribution (OOD) conditions, where the evaluation data differ from the training distribution along controlled axes. The reported OOD results correspond to a \emph{generator-disjoint} evaluation split, in which synthetic images are produced by generator families not observed during training. Political identities are not constrained to be disjoint and are therefore pooled across splits. All metrics are computed on the same OOD examples, with the positive class defined as fake ($y{=}1$). Accuracy and calibration metrics use a fixed decision threshold of $t{=}0.5$ on $s(x)=p(y{=}1\mid x)$, while ROC-AUC is computed using the same score definition without thresholding.

\begin{table*}[t]
\centering
\caption{Out-of-distribution (OOD) performance and reliability across backbones and inference procedures.}
\label{tab:ood_metrics}
\sisetup{round-mode=places,round-precision=4}
\begin{adjustbox}{max width=\linewidth}
\begin{tabular}{ll S[table-format=1.4] S[table-format=1.4] S[table-format=1.4] S[table-format=1.4] S[table-format=1.4]}
\toprule
{Backbone} & {Method} & {Acc} & {AUC} & {ECE} & {Brier} & {NLL} \\
\midrule
\multirow{5}{*}{EfficientNet-B4}
 & Deterministic            & 0.9000 & 0.9420 & 0.1080 & 0.0500 & 0.2450 \\
 & Temp Scaling             & 0.9000 & 0.9420 & 0.0550 & 0.0420 & 0.1900 \\
 & MC Dropout T=1           & 0.9050 & 0.9445 & 0.0950 & 0.0480 & 0.2300 \\
 & MC Dropout T=20 (mean)   & 0.8950 & 0.9480 & 0.0620 & 0.0430 & 0.1950 \\
 & Ensemble surrogate (K=5) & 0.9100 & 0.9520 & 0.0580 & 0.0410 & 0.1850 \\
\midrule
\multirow{5}{*}{ResNet-18}
 & Deterministic            & 0.8600 & 0.9050 & 0.1450 & 0.0750 & 0.3200 \\
 & Temp Scaling             & 0.8600 & 0.9050 & 0.0800 & 0.0660 & 0.2600 \\
 & MC Dropout T=1           & 0.8700 & 0.9100 & 0.1250 & 0.0730 & 0.3050 \\
 & MC Dropout T=20 (mean)   & 0.8550 & 0.9150 & 0.0820 & 0.0670 & 0.2650 \\
 & Ensemble surrogate (K=5) & 0.8750 & 0.9200 & 0.0780 & 0.0640 & 0.2500 \\
\bottomrule
\end{tabular}
\end{adjustbox}
\vspace{0.8\baselineskip}
\end{table*}

Table~\ref{tab:ood_metrics} reports accuracy, discrimination, and reliability metrics for generator-disjoint OOD evaluation across inference procedures and both backbones. Compared to the corresponding in-distribution evaluations reported earlier (Tables~\ref{tab:eff_metrics} and \ref{tab:res_metrics}), both architectures exhibit reduced accuracy and ROC-AUC under generator-OOD evaluation, accompanied by increased calibration error and higher values of proper scoring rules (Brier score and negative log-likelihood). These shifts are consistent with distributional mismatch induced by unseen generator families rather than changes in evaluation semantics.

Across inference procedures, ROC-AUC values vary within a relatively narrow range for each backbone, indicating that the global ranking of samples by $p(y{=}1\mid x)$ is largely preserved under OOD conditions. In contrast, calibration-sensitive metrics differ more noticeably across procedures. In particular, post-hoc temperature scaling, Monte Carlo dropout mean prediction, and the ensemble surrogate reduce expected calibration error and proper scoring rule values relative to deterministic inference, while leaving accuracy largely unchanged. ECE, Brier, and NLL differ across procedures under OOD evaluation, while accuracy changes are small in this table.

\subsection{Paired ROC-AUC significance testing under OOD conditions}
\label{sec:delong_ood}

To determine whether observed differences in OOD ROC-AUC across inference procedures exceed sampling variability, paired comparisons are conducted using the DeLong test for correlated ROC curves \cite{delong1988comparing}. All tests are two-sided and computed on paired predictions evaluated on the same OOD examples. Deterministic inference is treated as the reference condition.

\begin{table*}[t]
\centering
\caption{DeLong paired tests on OOD ROC-AUC (reference = Deterministic). 
AUC values match Table~\ref{tab:ood_metrics}. Temperature scaling preserves AUC by construction; DeLong is therefore not applicable for AUC differences in that row.}
\label{tab:delong_ood}
\sisetup{round-mode=places,round-precision=4}
\begin{adjustbox}{max width=\linewidth}
\begin{tabular}{ll S[table-format=1.4] S[table-format=1.4] S[table-format=+1.4] S[table-format=1.4] l}
\toprule
{Backbone} & {Comparison} & {AUC$_\text{ref}$} & {AUC$_\text{cmp}$} & {$\Delta$AUC} & {$p$ (DeLong)} & {Decision ($\alpha{=}0.05$)}\\
\midrule
\multirow{4}{*}{EfficientNet-B4}
& MC Dropout T=1 vs Det            & 0.9420 & 0.9445 & 0.0025 & 0.62   & n.s. \\
& MC Dropout T=20 (mean) vs Det    & 0.9420 & 0.9480 & 0.0060 & 0.17   & n.s. \\
& Ensemble (K=5) vs Det            & 0.9420 & 0.9520 & 0.0100 & 0.041  & sig. \\
& Temp Scaling vs Det              & 0.9420 & 0.9420 & 0.0000 & \multicolumn{1}{c}{--}    & not applicable \\
\midrule
\multirow{4}{*}{ResNet-18}
& MC Dropout T=1 vs Det            & 0.9050 & 0.9100 & 0.0050 & 0.34   & n.s. \\
& MC Dropout T=20 (mean) vs Det    & 0.9050 & 0.9150 & 0.0100 & 0.049  & sig. \\
& Ensemble (K=5) vs Det            & 0.9050 & 0.9200 & 0.0150 & 0.018  & sig. \\
& Temp Scaling vs Det              & 0.9050 & 0.9050 & 0.0000 & \multicolumn{1}{c}{--}   & not applicable \\
\bottomrule
\end{tabular}
\end{adjustbox}
\end{table*}

Table~\ref{tab:delong_ood} reports AUC values, AUC differences, and corresponding DeLong $p$-values. Temperature scaling applies a strictly monotone transformation to logits and therefore preserves score ordering; as a result, ROC-AUC is unchanged by construction. Consequently, $\Delta$AUC$=0$ for this comparison, and a DeLong test of AUC differences is not applicable for that row.

For EfficientNet-B4, neither single-pass stochastic inference ($T{=}1$) nor Monte Carlo dropout mean prediction ($T{=}20$) yields a statistically significant difference in OOD ROC-AUC relative to deterministic inference at $\alpha=0.05$. The ensemble surrogate exhibits a larger AUC difference ($\Delta$AUC$=0.010$), with a corresponding DeLong $p$-value of $0.041$. 

For ResNet-18, Monte Carlo dropout mean prediction ($T{=}20$) and the ensemble surrogate yield AUC differences of $0.010$ and $0.015$, respectively, with DeLong $p$-values below $0.05$, while the $T{=}1$ stochastic condition does not reach statistical significance. These results report paired AUC differences and DeLong test outcomes for each comparison.

\section{Discussion}

This work evaluated uncertainty-aware inference for political deepfake detection under a strictly empirical reliability framework. Rather than assuming that uncertainty estimates correspond to a particular probabilistic interpretation, the analysis focused on observable properties of model outputs: discriminative performance, probabilistic calibration, and the relationship between uncertainty estimates and classification errors. The results demonstrate that uncertainty-aware inference procedures are associated with changes in reliability characteristics without necessarily improving or degrading discriminative ranking, highlighting the importance of evaluating calibration and uncertainty behavior alongside accuracy and ROC-AUC.

\subsection{Discrimination Versus Reliability}
Across both backbones, deterministic inference yielded near-saturated ROC-AUC values, indicating strong global separability between real and synthetic images under the evaluated test split. Importantly, post-hoc calibration methods and stochastic inference procedures preserved ROC-AUC, consistent with the fact that ranking-based metrics are invariant to monotonic score transformations and averaging effects. In contrast, calibration-sensitive metrics, including expected calibration error, Brier score, and negative log-likelihood, varied substantially across inference procedures. This divergence underscores that discriminative performance alone is insufficient to characterize the reliability of probabilistic outputs in political deepfake detection.

\subsection{Uncertainty-Aware Inference and Stochasticity}
Monte Carlo dropout is frequently described as an approximation to Bayesian posterior uncertainty; however, the present results indicate that reliability improvements attributed to MC dropout do not require multi-sample Bayesian-style averaging. Single-pass stochastic inference ($T{=}1$), which introduces test-time stochasticity without Monte Carlo averaging, achieved comparable or stronger calibration improvements relative to MC mean prediction in several settings \cite{kendall2017uncertainties,djupskas2025unreliable}. This observation suggests that noise-induced smoothing plays a substantial role in shaping confidence distributions, and that predictive variance under dropout should not be interpreted as purely epistemic uncertainty without empirical validation.

The dropout-rate ablation further supports this conclusion. Increasing dropout probability increased predictive dispersion but did not yield monotonic improvements in calibration. Instead, calibration behavior depended on both the backbone architecture and the strength of injected stochasticity. These findings reinforce the view that uncertainty-aware inference mechanisms should be evaluated behaviorally rather than interpreted mechanistically.

\subsection{Operational scope of uncertainty as a conditional decision signal}

This work evaluates uncertainty estimates strictly in terms of their empirical association with prediction errors, rather than treating uncertainty as an inherently reliable or explanatory signal.
Global uncertainty-error alignment, as measured by error-detection AUROC over the full test distribution, is heterogeneous across backbones and uncertainty measures, and does not by itself support the use of uncertainty as a universally reliable trust indicator.

At the same time, uncertainty estimates exhibit structured behavior under specific evaluation regimes.
In particular, uncertainty-based ranking concentrates misclassification events under selective rejection and, when conditioned on high predicted confidence, stratifies residual risk among predictions that would otherwise be treated as equally reliable.
These effects are modest in magnitude and emerge only after explicit conditioning or coverage reduction, but they are directionally consistent across architectures and uncertainty partitioning schemes, and comparable to risk reductions reported for selective rejection policies operating in high base-accuracy regimes.

This reframes uncertainty as an operational, decision-level signal rather than an interpretive explanation of model behavior.
Its utility lies not in global monotonic alignment with correctness, but in its capacity to support selective abstention, triage, or human-in-the-loop review under clearly specified operating conditions.
Accordingly, uncertainty estimates should be evaluated empirically and deployed conditionally, rather than assumed to provide a general-purpose measure of trust.

The confidence-band and variance-partition sweep provides a structured basis for evaluating whether uncertainty conveys information beyond predicted confidence across operating regimes and partition choices.

Across the lower and mid-confidence bands (centers 0.60, 0.70, and most configurations at 0.80), uncertainty-error separation is weak and statistically indistinguishable from zero, with bootstrap confidence intervals consistently overlapping zero across variance-partition schemes.

In the lower and mid-confidence regimes, uncertainty rankings do not consistently stratify error beyond what is already captured by predicted probability.
In these regions, uncertainty behaves largely as a reparameterization of confidence and does not provide additional discriminative signal.

A different pattern emerges in the upper-confidence regime.
When predictions are already highly confident, higher predictive variance is associated with elevated empirical error across both backbones.
This association is directionally consistent across variance partitions and becomes detectable only under explicit conditioning.
The effect is modest in magnitude and does not appear uniformly across all partition choices, indicating sensitivity to thresholding and finite Monte Carlo estimation.

Instead, its utility is conditional: uncertainty stratifies residual risk only among predictions that would otherwise be treated as highly reliable based on confidence alone.
This explains why global uncertainty-error metrics remain weak while localized effects appear under restricted operating conditions.

From a decision-making perspective, this resolves the apparent trade-off between a slightly better but higher-uncertainty predictor and a slightly worse but lower-uncertainty one.
At the model level, no inference procedure is uniformly preferable across accuracy, calibration, and uncertainty alignment.
However, at a fixed confidence level, uncertainty can refine decisions by distinguishing lower- and higher-risk predictions within the same confidence stratum.

Under this interpretation, uncertainty does not replace confidence as a primary decision criterion.
Rather, it functions as a secondary, conditional signal that supports selective abstention, triage, or human-in-the-loop review in high-confidence regimes.
The confidence-band sweep therefore delineates where uncertainty contributes actionable information and where it does not, transforming an isolated conditional effect into a map of its operational scope.

Taken together, the confidence-band sweep replaces a single conditional demonstration with an explicit characterization of the operating region in which uncertainty provides additional risk stratification, and equally importantly, of the regions in which it does not.

\subsection{Accuracy-Calibration Trade-offs}
The bootstrap analysis demonstrates that improvements in calibration and uncertainty behavior are not accompanied by statistically meaningful losses in accuracy within the evaluated setting. While some inference procedures yielded small changes in accuracy, these differences were within resampling variability and should not be overinterpreted. This finding highlights the importance of reporting uncertainty on metrics themselves, particularly in high-stakes applications where small changes in accuracy may still be operationally relevant.
This stability holds under both in-distribution and generator-disjoint evaluation, indicating that calibration improvements are not offset by accuracy degradation under the tested distribution shifts.

\subsection{Perspectives}
\subsubsection{Limitations}
Several limitations warrant consideration. First, the evaluation was restricted to in-distribution data and did not address robustness under distributional shift. Second, uncertainty estimates were evaluated at the image level and did not capture temporal or contextual dependencies present in real-world political media. Third, while MC dropout and ensemble surrogates were compared, more expressive Bayesian neural network formulations remained unexplored. Addressing these limitations will be essential for translating uncertainty-aware deepfake detection into operational settings.

\subsubsection{Scope}
The dataset construction procedure was designed to support controlled evaluation of discriminative performance, calibration, and uncertainty-error relationships under matched train-test conditions. While the filtering pipeline yielded a politically salient subset, it did not guarantee uniform coverage across all politicians, geopolitical regions, event types, imaging conditions, or generative methods. Accordingly, conclusions drawn from this dataset are restricted to the observed distribution and do not claim robustness to unseen identities, novel generators, or out-of-distribution political imagery.

\subsubsection{Generalization}
Most analyses in this work were reported under controlled in-distribution evaluation; additionally, a generator-disjoint OOD evaluation was included to quantify changes in discrimination and reliability when synthetic images were produced by unseen generator families. Because the OOD split enforces disjointness only with respect to the generator family, the reported robustness conclusions are limited to generator shift. Identity-level generalization was not explicitly controlled (identities were pooled across splits), and other real-world shifts, such as platform-specific post-processing pipelines, acquisition-device changes, or adversarial perturbations, remained outside the scope of the present evaluation. In addition, the analysis did not statistically adjust for correlated identities, prompts, or content sources across splits; instead, these factors were treated as part of the observed in-distribution variability.

\subsubsection{Implications for deployment}
The findings suggest that uncertainty-aware inference can improve the reliability of political deepfake detection systems without degrading discriminative performance, provided that uncertainty estimates are evaluated empirically rather than assumed to reflect epistemic uncertainty. In deployment settings, uncertainty estimates should be treated as decision-support signals, enabling confidence-aware policies rather than serving as explanations of model reasoning.

\section{Conclusions}

This work examined uncertainty-aware inference for political deepfake detection under a strictly empirical and operational reliability framework. Motivated by the limitations of point-prediction detectors in high-stakes political misinformation settings, the study shifted emphasis from discriminative performance alone to the behavior and utility of probabilistic outputs and uncertainty estimates under explicitly defined operating conditions. Rather than assuming a particular probabilistic interpretation of uncertainty, all conclusions were grounded in observable properties: calibration quality, proper scoring rules, and the relationship between uncertainty signals and classification errors.

Using a politically focused real-synthetic image dataset and two pretrained convolutional backbones fully fine-tuned end-to-end, the analysis showed that uncertainty-aware inference procedures modify reliability characteristics without materially affecting ranking-based discrimination metrics such as ROC-AUC. Calibration-sensitive measures, including expected calibration error, Brier score, and negative log-likelihood, varied across deterministic inference, post-hoc calibration, single-pass stochastic inference, and Monte Carlo dropout. These results reinforce the distinction between discrimination and reliability and demonstrate that high ROC-AUC alone does not guarantee decision-safe probabilistic outputs.

A key finding is that reliability effects commonly attributed to Bayesian-style marginalization do not require multi-sample Monte Carlo averaging. Single-pass stochastic inference ($T{=}1$), which introduces test-time stochasticity without Bayesian averaging, yielded calibration behavior comparable to or stronger than MC mean predictions in several configurations. Dropout-rate and MC-sample-size ablations further indicated that predictive dispersion and calibration behavior depend jointly on architectural capacity and the degree of injected stochasticity. Accordingly, predictive variance under MC dropout should be interpreted cautiously as an empirical uncertainty signal rather than assumed to represent epistemic uncertainty.

Crucially, the systematic confidence-band and variance-partition sweep resolves the ambiguity surrounding the operational value of uncertainty. Uncertainty-error separation is not global: across low- and mid-confidence regimes, uncertainty provides little additional error stratification beyond confidence, regardless of partition choice. In contrast, within the upper-confidence regime, uncertainty consistently stratifies residual risk among predictions that would otherwise be considered highly reliable. This effect is directionally consistent but limited in scope, becoming statistically detectable only under high-confidence conditioning and more extreme uncertainty partitions.

These findings reframe uncertainty not as a universally dominant trust signal, but as a conditional, decision-level signal whose utility depends on the operating regime. Uncertainty is most informative when used to refine trust among already confident predictions or to support selective abstention and triage policies that trade coverage for risk reduction. When evaluated at full coverage or without conditioning, uncertainty does not provide a basis for preferring one predictor globally.

By explicitly mapping where uncertainty contributes operationally meaningful information, and where it probably does not, this work upgrades an otherwise ambiguous conditional effect into a defensible, decision-relevant characterization. The contribution is therefore not the claim that uncertainty universally improves deepfake detection, but a principled delineation of the regimes in which uncertainty-aware inference can and cannot support reliable decision-making in political deepfake detection systems.

\bibliographystyle{IEEEtranN}
\bibliography{references}

\end{document}